\documentclass{article}

\usepackage{microtype}
\usepackage{graphicx}
\usepackage{booktabs} 

\usepackage{amsfonts}
\usepackage{amsmath}

\usepackage{mathtools}
\usepackage{mleftright}
\usepackage{dsfont}
\usepackage{amscd,amssymb,amsmath,amsthm,bbold}
\usepackage{bm}
\usepackage[loading]{tracefnt}

\usepackage{subfig}

\newcommand{\ws}{\theta}
\newcommand{\ps}{\omega}

\usepackage{xcolor}
\definecolor{mypink}{rgb}{0.9,0.4,0.5}
\definecolor{myOrange}{rgb}{0.1,0.5,0.1}
\definecolor{purple}{rgb}{0.56862745098,0.25098039215
,0.63921568627}

\usepackage{hyperref}

\usepackage[accepted]{icml2021}

\icmltitlerunning{Learning Neural Network Subspaces}

\begin{document}
\newtheorem{theorem}{Theorem}[section]
\newtheorem{corollary}{Corollary}[theorem]
\newtheorem{lemma}[theorem]{Lemma}
\newtheorem{proposition}[theorem]{Proposition}

\newtheorem{definition}[theorem]{Definition}
\newtheorem{remark}[theorem]{Remark}

\newcommand{\round}[1]{\left( #1 \right)}
\newcommand{\curly}[1]{\left\lbrace #1 \right\rbrace}
\newcommand{\squarebrack}[1]{\left\lbrack #1 \right\rbrack}

\newcommand{\sumi}[2]{\sum\limits_{i=#1}^{#2}}
\newcommand{\sumj}[2]{\sum\limits_{j=#1}^{#2}}
\newcommand{\sumk}[2]{\sum\limits_{k=#1}^{#2}}
\newcommand{\sump}[2]{\sum\limits_{p=#1}^{#2}}
\newcommand{\suml}[2]{\sum\limits_{l=#1}^{#2}}
\newcommand{\sumn}[2]{\sum\limits_{n=#1}^{#2}}
\newcommand{\summ}[2]{\sum\limits_{m=#1}^{#2}}
\newcommand{\sumt}[2]{\sum\limits_{t=#1}^{#2}}

\newcommand{\Sum}{\sum_{i = 1}^{n}}
\newcommand{\Sumi}[1]{\sum\limits_{i = 1}^{#1}}
\newcommand{\Sumt}[1]{\sum\limits_{t = 1}^{#1}}

\newcommand{\abs}[1]{\left\lvert #1 \right\rvert}
\newcommand{\norm}[2]{\left\lVert#2\right\rVert_{#1}}
\newcommand{\esqnorm}[1]{\left\lVert#1\right\rVert_2^2}
\newcommand{\enorm}[1]{\left\lVert#1\right\rVert_2}
\newcommand{\infnorm}[1]{\left\lVert#1\right\rVert_\infty}
\newcommand{\opnorm}[1]{\left\lVert#1\right\rVert_\text{op}}
\newcommand{\normF}[1]{\left\lVert#1\right\rVert_{\text{F}}}
\newcommand{\inner}[1]{\left\langle#1\right\rangle}
\newcommand{\ceil}[1]{\left\lceil#1\right\rceil}
\newcommand{\floor}[1]{\left\lfloor#1\right\rfloor}

\newcommand{\zero}{\mathbf{0}}
\newcommand{\one}{\mathbf{1}}

\newcommand{\avec}{\mathbf{a}}
\newcommand{\bvec}{\mathbf{b}}
\newcommand{\cvec}{\mathbf{c}}
\newcommand{\dvec}{\mathbf{d}}
\newcommand{\e}{\mathbf{e}}
\newcommand{\f}{\mathbf{f}}
\newcommand{\g}{\mathbf{g}}
\newcommand{\h}{\mathbf{h}}
\newcommand{\ivec}{\mathbf{i}}
\newcommand{\jvec}{\mathbf{j}}
\newcommand{\kvec}{\mathbf{k}}
\newcommand{\lvec}{\mathbf{l}}
\newcommand{\m}{\mathbf{m}}
\newcommand{\n}{\mathbf{n}}
\newcommand{\ovec}{\mathbf{o}}
\newcommand{\p}{\mathbf{p}}
\newcommand{\q}{\mathbf{q}}
\newcommand{\rvec}{\mathbf{r}}
\newcommand{\s}{\mathbf{s}}
\newcommand{\tvec}{\mathbf{t}}
\newcommand{\uvec}{\mathbf{u}}
\newcommand{\vvec}{\mathbf{v}}
\newcommand{\w}{\mathbf{w}}
\newcommand{\x}{\mathbf{x}}
\newcommand{\y}{\mathbf{y}}
\newcommand{\z}{\mathbf{z}}

\newcommand{\A}{\mathbf{A}}
\newcommand{\B}{\mathbf{B}}
\newcommand{\C}{\mathbf{C}}
\newcommand{\D}{\mathbf{D}}
\newcommand{\Emat}{\mathbf{E}}
\newcommand{\F}{\mathbf{F}}
\newcommand{\G}{\mathbf{G}}
\newcommand{\Hmat}{\mathbf{H}}
\newcommand{\I}{\mathbf{I}}
\newcommand{\J}{\mathbf{J}}
\newcommand{\K}{\mathbf{K}}
\newcommand{\Lmat}{\mathbf{L}}
\newcommand{\M}{\mathbf{M}}
\newcommand{\N}{\mathbf{N}}
\newcommand{\Omat}{\mathbf{O}}
\newcommand{\Pmat}{\mathbf{P}}
\newcommand{\Q}{\mathbf{Q}}
\newcommand{\Rmat}{\mathbf{R}}
\newcommand{\Smat}{\mathbf{S}}
\newcommand{\T}{\mathbf{T}}
\newcommand{\U}{\mathbf{U}}
\newcommand{\V}{\mathbf{V}}
\newcommand{\W}{\mathbf{W}}
\newcommand{\X}{\mathbf{X}}
\newcommand{\Y}{\mathbf{Y}}
\newcommand{\Z}{\mathbf{Z}}

\newcommand{\SIGMA}{\mathbf{\Sigma}}
\newcommand{\LAMBDA}{\mathbf{\Lambda}}

\newcommand{\Acal}{\mathcal{A}}
\newcommand{\Bcal}{\mathcal{B}}
\newcommand{\Ccal}{\mathcal{C}}
\newcommand{\Dcal}{\mathcal{D}}
\newcommand{\Ecal}{\mathcal{E}}
\newcommand{\Fcal}{\mathcal{F}}
\newcommand{\Gcal}{\mathcal{G}}
\newcommand{\Hcal}{\mathcal{H}}
\newcommand{\Ical}{\mathcal{I}}
\newcommand{\Jcal}{\mathcal{J}}
\newcommand{\Kcal}{\mathcal{K}}
\newcommand{\Lcal}{\mathcal{L}}
\newcommand{\Mcal}{\mathcal{M}}
\newcommand{\Ncal}{\mathcal{N}}
\newcommand{\Ocal}{\mathcal{O}}
\newcommand{\Pcal}{\mathcal{P}}
\newcommand{\Qcal}{\mathcal{Q}}
\newcommand{\Rcal}{\mathcal{R}}
\newcommand{\Scal}{\mathcal{S}}
\newcommand{\Tcal}{\mathcal{T}}
\newcommand{\Ucal}{\mathcal{U}}
\newcommand{\Vcal}{\mathcal{V}}
\newcommand{\Wcal}{\mathcal{W}}
\newcommand{\Xcal}{\mathcal{X}}
\newcommand{\Ycal}{\mathcal{Y}}
\newcommand{\Zcal}{\mathcal{Z}}

\newcommand{\alphavec}{\boldsymbol{\alpha}}
\newcommand{\betavec}{\boldsymbol{\beta}}
\newcommand{\gammavec}{\boldsymbol{\gamma}}
\newcommand{\deltavec}{\boldsymbol{\delta}}
\newcommand{\epsvec}{\boldsymbol{\epsilon}}
\newcommand{\etavec}{\boldsymbol{\eta}}
\newcommand{\nuvec}{\boldsymbol{\nu}}
\newcommand{\tauvec}{\boldsymbol{\tau}}
\newcommand{\rhovec}{\boldsymbol{\rho}}
\newcommand{\lmbda}{\boldsymbol{\lambda}}
\newcommand{\muvec}{\boldsymbol{\mu}}
\newcommand{\thetavec}{\boldsymbol{\theta}}

\newcommand{\BigO}[1]{\mathcal{O}\round{#1}}
\newcommand{\BigOmega}[1]{\Omega\round{#1}}

\newcommand{\R}{\mathbb{R}}
\newcommand{\Rd}[1]{\mathbb{R}^{#1}}
\newcommand{\Natural}{\mathbb{N}}
\newcommand{\Complex}{\mathbb{C}}
\newcommand{\Integer}{\mathbb{Z}}
\newcommand{\Rational}{\mathbb{Q}}

\newcommand{\E}[1]{\mathbb{E}\squarebrack{#1}}
\newcommand{\Exp}[2]{\mathbb{E}_{#1}\squarebrack{#2}}
\newcommand{\Prob}[1]{\mathds{P}\squarebrack{#1}}
\newcommand{\Probability}[2]{P_{#1}\curly{#2}}
\newcommand{\Var}[1]{\mathrm{Var}\squarebrack{#1}}
\newcommand{\Cov}[1]{\mathrm{Cov}\squarebrack{#1}}
\newcommand{\PR}[1]{\mathds{P}\round{#1}}

\newcommand{\inv}[1]{\frac{1}{#1}}
\newcommand{\indicator}[2]{\mathbbm{1}_{#1}\squarebrack{#2}}
\newcommand{\Tr}[1]{\text{Tr}\squarebrack{#1}}

\newcommand{\BOX}[1]{\fbox{\parbox{\linewidth}{\centering#1}}}
\newcommand{\textequal}[1]{\stackrel{#1}{=}}
\newcommand{\textleq}[1]{\stackrel{#1}{\leq}}
\newcommand{\textgeq}[1]{\stackrel{#1}{\geq}}
\newcommand{\defeq}{\vcentcolon=}

\newcommand{\dd}[2]{\frac{d #1}{d #2}}
\newcommand{\ddn}[3]{\frac{d^{#1} #2}{d #3^{#1}}}
\newcommand{\dodo}[2]{\frac{\partial #1}{\partial #2}}
\newcommand{\dodon}[3]{\frac{\partial^{#1} #2}{\partial {#3}^{#1}}}

\twocolumn[
\icmltitle{Learning Neural Network Subspaces}



\icmlsetsymbol{equal}{*}

\begin{icmlauthorlist}
\icmlauthor{Mitchell Wortsman}{to}
\icmlauthor{Maxwell Horton}{goo}
\icmlauthor{Carlos Guestrin}{goo}
\icmlauthor{Ali Farhadi}{goo}
\icmlauthor{Mohammad Rastegari}{goo}
\end{icmlauthorlist}

\icmlaffiliation{to}{University of Washington (work completed during internship at Apple).}
\icmlaffiliation{goo}{Apple}

\icmlcorrespondingauthor{Mitchell Wortsman}{mitchnw@cs.washington.edu}

\icmlkeywords{Machine Learning, ICML}

\vskip 0.3in
]



\printAffiliationsAndNotice{}

\begin{abstract}
Recent observations have advanced our understanding of the neural network optimization landscape, revealing the existence of (1) paths of high accuracy containing diverse solutions and (2) wider minima offering improved performance. Previous methods observing diverse paths require multiple training runs. In contrast we aim to leverage both property (1) and (2) with a single method and in a single training run. With a similar computational cost as training one model, we learn lines, curves, and simplexes of high-accuracy neural networks. These neural network subspaces contain diverse solutions that can be ensembled, approaching the ensemble performance of independently trained networks without the training cost. Moreover, using the subspace midpoint boosts accuracy, calibration, and robustness to label noise, outperforming Stochastic Weight Averaging.
\end{abstract}

\section{Introduction}
\label{sec:introduction}

Optimizing a neural network is often conceptualized as finding a minimum in an objective landscape. Therefore, understanding the geometric properties of this landscape has emerged as an important goal. Recent work has illuminated many intriguing phenomena.
\citet{garipov2018loss, draxler2018essentially} determine that independently trained models are connected by a curve in weight space along which loss remains low. Additionally, \citet{frankle2020linear} demonstrate that networks which share only a few epochs of their optimization trajectory are connected by a linear path of high accuracy. However, the connected regions in weight space found by \citet{garipov2018loss, draxler2018essentially, frankle2020linear} require approximately twice the training time compared with standard training, as two separate minima are first identified then connected.

\begin{figure}[!htbp]
    \centering
    \includegraphics[width=\columnwidth]{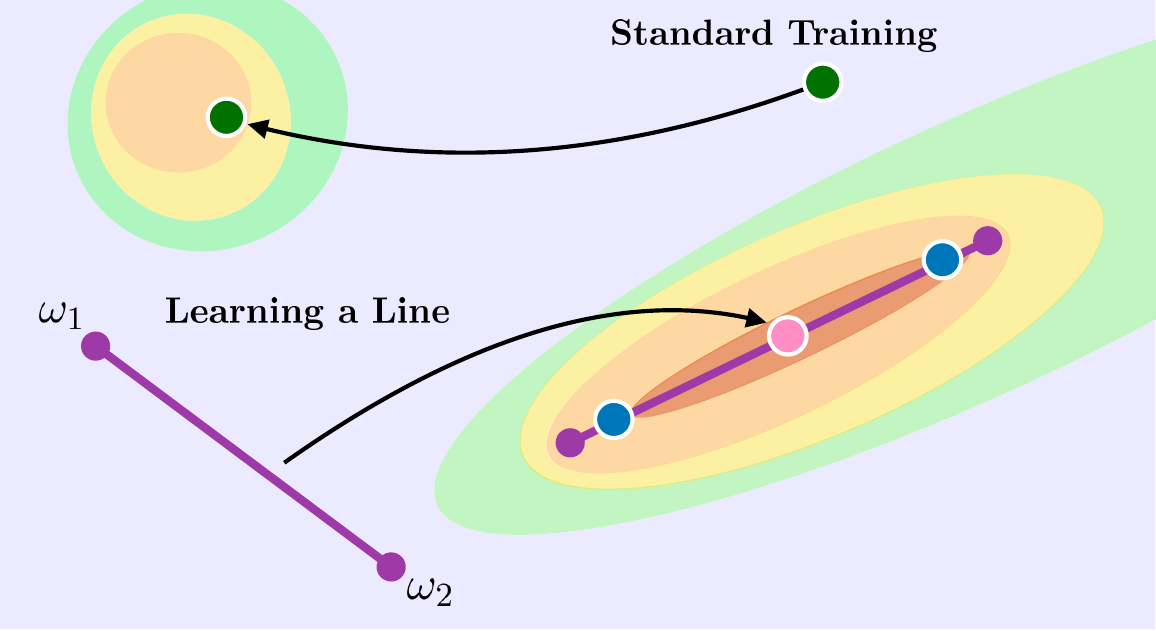}
    \caption{Schematic for {\color{purple}learning a line} of neural networks compared with {\color{myOrange}standard training}. The {\color{mypink} midpoint} outperforms standard training in terms of accuracy, calibration, and robustness. {\color{blue}Models near the endpoints} enable high-accuracy ensembles in a single training run. }
    \vspace{-1.5em}
    \label{fig:teaser}
\end{figure}

This work is motivated by the existence of connected, functionally diverse regions in solution space. In contrast to prior work, our aim is to directly parameterize and learn these neural network subspaces from scratch in a single training run. For instance, when training a line (\autoref{fig:teaser}) we begin with two randomly initialized endpoints and consider the neural networks on the linear path which connects them. At each iteration we use a randomly sampled network from the line, backpropagating the training loss to update the endpoints. Central to our method is a regularization term which encourages orthogonality between the endpoints, just as two independently trained networks are orthogonal \cite{fort2019deep}. When the line settles into a low loss region we find that models from opposing ends are functionally diverse.

\begin{figure*}[!htbp]
  \centering
  \subfloat{\includegraphics[width=0.33\textwidth]{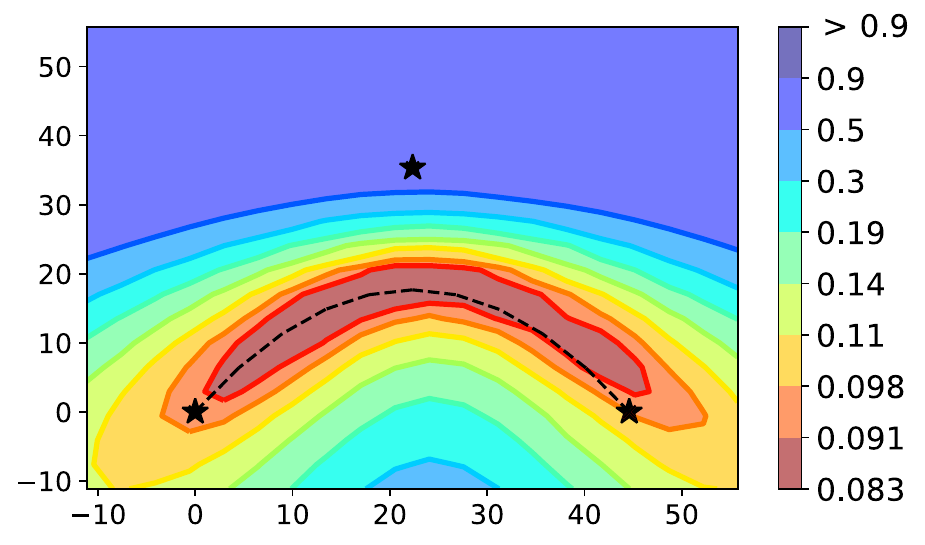}\label{fig:vizcurve}}
  \hfill
  \subfloat{\includegraphics[width=0.33\textwidth]{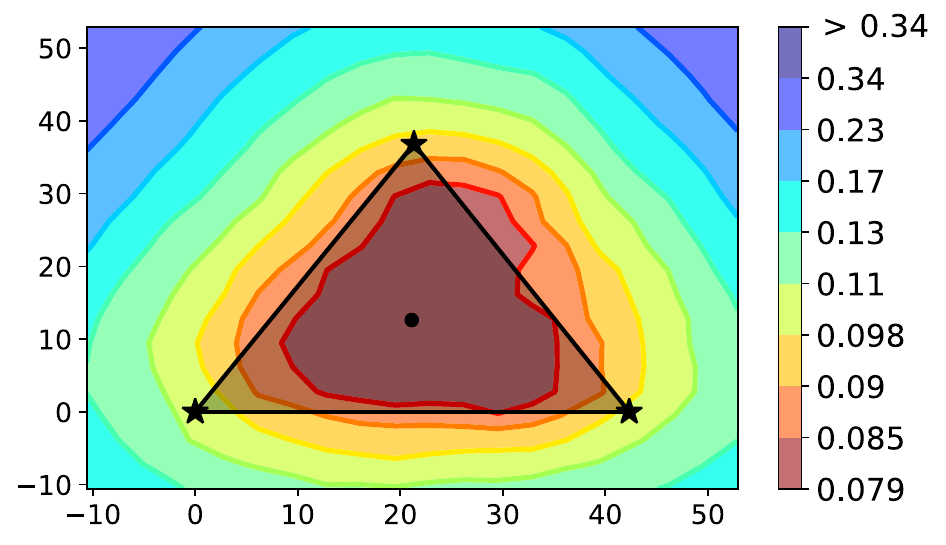}\label{fig:vizplane}}
    \hfill
  \subfloat{\includegraphics[width=0.33\textwidth]{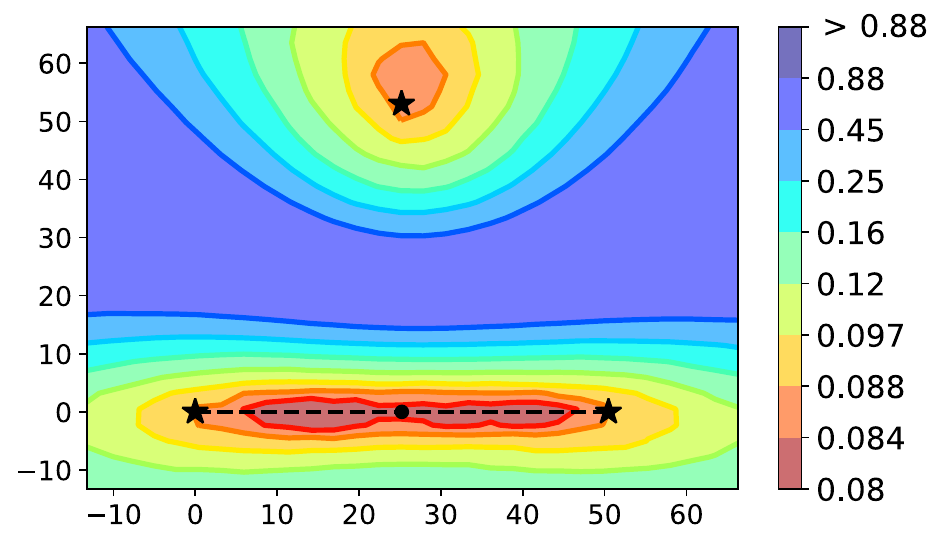}\label{fig:vizline}}
    \vspace*{-1em}
  \caption{Test error on a two dimensional plane for three learned subspaces for cResNet20 (CIFAR10)---a quadratic Bezier curve (left), a simplex with three endpoints (middle), and a line (right). The subspace parameters $\omega_1$, $\omega_2$ and $\omega_3$ are plotted and used to construct the plane, except for the line for which $\omega_3$ was taken to be a solution obtained via standard training. Note that although $\omega_3$ is used to define the Bezier curve (left), it never passes through it. 
  Visualization as in \citet{garipov2018loss} with $\omega_1$ at the origin.}
    \label{fig:planes}
  \vspace*{-1em}
\end{figure*}

In addition to lines, we learn curves and simplexes of high-accuracy neural networks (\autoref{fig:planes}). We also uncover benefits beyond functional diversity. Lines and simplexes identify and traverse large flat minima, with endpoints near the periphery. The midpoint corresponds to a less sharp solution, which is associated with better generalization \cite{pmlr-v80-dziugaite18a}. Using this midpoint corresponds to ensembling in weight space, producing a single model which requires no additional compute during inference. We find that taking the midpoint of a simplex can boost accuracy, calibration, and robustness to label noise. 


The rest of the paper is organized via the following contributions:
\begin{enumerate}
    \item We contextualize our work via 5 observations regarding the objective landscape (\autoref{sec:prelim}).
    \item We introduce a method for learning diverse and high-accuracy lines, curves, and simplexes of neural networks (\autoref{sec:meth}).
    \item We show that lines and curves found in a single training run contain models that approach or match the ensemble accuracy of independently trained networks  (\autoref{sec:reslines}).
    \item We find that taking the midpoint of a simplex provides a boost in accuracy, calibration, and robustness (\autoref{sec:simplex}; \autoref{sec:large}).
\end{enumerate}

\section{Preliminaries and Related Methods}
\label{sec:prelim}

We highlight a few recent observations which have advanced understanding of the neural network optimization landscape  \cite{NIPS2014_17e23e50, li2018measuring,li2018visualizing,fort2019large, evci2019difficulty, frankle2020revisiting, oswald2021neural}.
We remain in the setting of image classification with setup and notation drawn from \citet{frankle2020linear}.

Consider a neural network $f\mleft(\x, \ws \mright)$ with input $\x$ and parameters $\ws \in \R^n$. For initial random weights $\ws_0$ and SGD randomness $\xi$, the weights at epoch $t$ are given by $\ws_t =
\mathsf{Train}^{0 \rightarrow t}\mleft(\ws_0, \xi \mright)$. Additionally let $\mathsf{Acc}\mleft( \ws \mright)$ denote the test accuracy of network $f$ with parameters $\ws$.
The first three observations pertain to the setting where two networks are trained with different SGD noise---consider $\ws^1_T = \mathsf{Train}^{0 \rightarrow T}\mleft(\ws_0, \xi_1 \mright)$ and  $\ws^2_T = \mathsf{Train}^{0 \rightarrow 
T}\mleft(\ws_0, \xi_2 \mright)$. The observations are unchanged when $\theta_T^1$ and $\theta_T^2$ have differing initializations.

\textbf{Observation 1.} \citep{lakshminarayanan2017simple} Ensmembling $\ws^1_T$ and $\ws^2_T$ in output space---making predictions $\hat\y = \frac{1}{2}\mleft( f\mleft(\x, \ws_T^1\mright) + f\mleft(\x, \ws_T^2\mright)\mright)$---boosts accuracy, calibration, and robustness. This is attributed to functional diversity meaning $f\mleft(\cdot, \ws^1_T\mright)$ and $f\mleft(\cdot, \ws^2_T\mright)$ make different errors.

\textbf{Observation 2.} \citep{frankle2020linear, fort2020deep} Ensmembling $\ws^1_T$ and $\ws^2_T$ in weight space---making predictions with the network $f\mleft(\x, \frac{1}{2} \mleft(\ws_T^1 + \ws_T^2 \mright) \mright)$---fails, achieving no better accuracy than an untrained network.

\textbf{Definition 1.}
A \emph{connector} between neural network weights $\psi_1,\psi_2 \in \R^n$ is a continuous function $\mathsf{P} : [0,1] \rightarrow \R^n$  such that $\mathsf{P}(0) = \psi_1$, $\mathsf{P}(1) = \psi_2$, and the worst-case accuracy along the connector is at least the average accuracy given by the weights at the endpoints. Equivalently, $\inf_{\alpha \in [0,1]}{ \mathsf{Acc}\mleft( \mathsf{P}(\alpha) \mright)} \gtrapprox \frac{1}{2}\mleft(\mathsf{Acc}\mleft(\psi_1\mright) + \mathsf{Acc}\mleft(\psi_2\mright) \mright)$. In the language of \textit{connectors}, Observation 2 states that there does not exist a \emph{linear} connector between $\ws_T^1$ and $\ws_T^2$. 

\textbf{Observation 3.} \citep{garipov2018loss,draxler2018essentially} There exists a \emph{nonlinear} connector $\mathsf{P}$ between $\ws_1^T$ and $\ws_2^T$, for instance a quadratic Bezier curve.

\textbf{Observation 4.} \citep{frankle2020linear} There exists a \emph{linear} connector when part of the optimization trajectory is shared. Instead of branching off at $\ws_0$, let $\ws_k = \mathsf{Train}^{0 \rightarrow k}\mleft(\ws_0, \xi \mright)$ and consider $\ws^i_{k\rightarrow T} = \mathsf{Train}^{k \rightarrow T}\mleft(\ws_k, \xi_i \mright)$ for $i \in \{1,2\}$. For $k \ll T$,  $\mathsf{P(\alpha)} = (1-\alpha) \ws^1_{k\rightarrow T} + \alpha \ws^2_{k\rightarrow T}$ is a linear connector. 

Observation 4 generalizes to the higher dimensional case (\autoref{sec:morefrankle}) for which a convex hull of neural networks attains high accuracy. To consider higher dimensional connectors we discuss one additional definition. Let $\Delta^{m-1} = \mleft\{ \bm{\alpha} \in \R^m : \sum_i \bm{\alpha}_i = 1, \bm{\alpha}_i \geq 0 \mright\}$ and let $\e_i$ refer to the standard basis vector (all zeros except for position $i$ which is $1$). Note that $\Delta^{m-1}$ is often referred to as the $m-1$ dimensional probability simplex.

\textbf{Definition 2.} An $m$-\textit{connector} on $\psi_1,...,\psi_m \in \R^n$ is a continuous function $\mathsf{P} : \Delta^{m-1} \rightarrow \R^n$ such that $\mathsf{P}(\e_i) = \psi_i$ and $\inf_{\bm{\alpha} \in \Delta^{m-1}}{\mathsf{Acc}\mleft( \mathsf{P}(\bm{\alpha}) \mright)} \gtrapprox \frac{1}{m}\sum_{i=1}^m \mathsf{Acc}\mleft( \psi_i \mright)$. This definition formalizes that in \citet{fort2019large}. In this work we will primarily focus on linear $m$-connectors which have the form $\mathsf{P}(\bm{\alpha}) = \sum_i \bm{\alpha}_i \psi_i$.

Linear $m$-connectors are implicitly used by \citet{izmailov2018averaging} in \emph{Stochastic Weight Averaging (SWA)}. SWA uses a high constant (or cyclic) learning rate towards the end of training to bounce around a minimum while occasionally saving checkpoints. SWA returns the weight space ensemble (average) of these models, motivated by the observation that SGD solutions often lie at the edge of a minimum and averaging moves towards the center. The averaged solution is less sharp, which may lead to better generalization \citep{chaudhari2019entropy, pmlr-v80-dziugaite18a, foret2020sharpness}. 

\textbf{Observation 5.} \citep{izmailov2018averaging} If weights $\psi_1$,...$\psi_m$ lie at the periphery of wide and flat low loss region, then $ \mathsf{Acc}\mleft( \frac{1}{m}\sum_{i=1}^m \psi_i \mright) > \frac{1}{m}\sum_{i=1}^m \mathsf{Acc}\mleft( \psi_i \mright)$.

SWA is extended by SWA-Gaussian
\citep{maddox2019simple} (which fits a Gaussian to the saved checkpoints) and \citet{izmailov2020subspace} (who considers the subspace which they span). These techniques advance Bayesian deep learning---methods which aim to learn a distribution over the parameters. Other Bayesian apporaches include variational methods \cite{blundell2015weight}, MC-dropout \cite{gal2016dropout}, and MCMC methods \cite{welling2011bayesian, Zhang2020Cyclical}. However, variational methods tend not to scale to larger networks such as residual networks \cite{maddox2019simple}. Moreover, a detailed empirical study by \citet{fort2019deep} recently observed that many Bayesian models tend to capture the local uncertainty of a single mode but are much less functionally diverse than independently trained networks which identify multiple modes. Ensembling models sampled from the learned distribution is therefore inferior in terms of accuracy and robustness. 

Other related techniques include \textit{Snapshot Ensembles (SSE)} \cite{huang2017snapshot} which use a cyclical learning rate with multiple restarts, saving checkpoints prior to each restart. Fast Geometric Ensembles \cite{garipov2018loss} employs a similar strategy but does not begin saving checkpoints until later in training. Other methods to efficiently train and evaluate ensembles include BatchE \cite{wen2020batchensemble}. Although their method is compelling, BatchE requires longer training for ensemble members to match standard training accuracy. 

To summarize, connectors---high-accuracy subspaces of neural networks---have two useful properties:
\begin{itemize}
    \item Property 1: They contain models which are functionally diverse and may be ensembled in output space (Observations 1 \& 3).
    \item Property 2: Taking the midpoint of the subspace (ensembling in weight space) can improve accuracy and generalization (Observation 5).
\end{itemize}
Prior work satisfying Property 1 requires multiple training runs. Subspaces satisfying Property 2 yield solutions that are less functionally diverse \cite{fort2019deep}. 
Our aim is to leverage both Property 1 and 2 in a single training run.


\section{Method} \label{sec:meth}

In a single training run, we find a connected region in solution space comprised of high-accuracy and diverse neural networks. To do so we directly parameterize and learn the parameters of a subspace.

First consider learning a line. Recall that the line between 
$\omega_1 \in \R^n$ and $\omega_2 \in \R^n$ in weight space is
$\mathsf{P}(\alpha; \omega_1, \omega_2) = (1-\alpha)\omega_1 + \alpha\omega_2$ for $\alpha$ in the domain $\Lambda = [0,1]$. Our goal is to learn parameters $\omega_1, \omega_2$ such that $\mathsf{Acc}\mleft(\mathsf{P}(\alpha; \omega_1, \omega_2)\mright)$ is high for all values of $\alpha \in \Lambda$ ($\mathsf{Acc}\mleft(\theta\mright)$ denotes the test accuracy of the neural network $f$ with weights $\theta$). Equivalently, our aim is to learn a high-accuracy connector between $\omega_1$ and $\omega_2$ (Definition 1).

\begin{algorithm}[t]
   \caption{$\mathsf{TrainSubspace}$}
   \label{alg:example}
\begin{algorithmic}
   \STATE {\bfseries Input:} $\mathsf{P}$ with domain $\Lambda$ and parameters $\mleft\{ \ps_i \mright\}_{i=1}^m$, network $f$, train set $\mathcal{S}$, loss $\ell$, and scalar $\beta$ (\textit{e.g.} a line has $\Lambda = [0,1]$ and $\mathsf{P}(\alpha; \ps_1, \ps_2) = (1-\alpha) \ps_1 + \alpha \ps_2$).
   \STATE{Initialize each $\ps_i$ independently.}
   \FOR{batch $(\x, \y) \subseteq \mathcal{S}$}
   \STATE{Sample $\bm{\alpha}$ uniformly from $\Lambda$}.
   \STATE{$\ws \gets \mathsf{P}\mleft(\bm{\alpha}; \mleft\{ \ps_i \mright\}_{i=1}^m \mright)$}
   \STATE{$\hat \y \gets f\mleft(\x, \ws \mright)$}
   \STATE{Sample $j, k$ from $\{1,...,m\}$ without replacement.}
   \STATE{$\mathcal{L} \gets \ell\mleft(\hat{\y}, \y\mright) + \beta \cos^2\mleft(\ps_j, {\ps_k} \mright)$}
   \STATE{Backprop $\mathcal{L}$ to each $\ps_i$ and update with SGD \& momentum using estimate $\frac{\partial \mathcal{L}}{\partial \omega_i} = 
   \frac{\partial \ell}{\partial \theta}\frac{\partial \mathsf{P}}{\partial \omega_i} + \beta \frac{\partial \cos^2\mleft(\ps_j, {\ps_k} \mright)}{\partial \omega_i}$}.
   \ENDFOR
\end{algorithmic}
\end{algorithm}
More generally we consider subspaces defined by $\mathsf{P}\mleft(\cdot, \{\omega_i\}_{i=1}^m \mright) : \Lambda \rightarrow \R^n$.
We experiment with two shapes in addition to lines:
\begin{enumerate}
    \item One-dimensional Bezier curves with a single bend $\mathsf{P}\mleft(\alpha; \omega_1, \omega_2, \omega_3 \mright) = (1-\alpha)^2 \omega_1 + 2\alpha(1-\alpha)\omega_3 + \alpha^2 \omega_2$ for $\alpha \in \Lambda = [0,1]$.
    \item Simplexes with $m$ endpoints $\{\omega_i\}_{i=1}^m$. A simplex is the convex hull defined by $\mathsf{P}\mleft(\bm{\alpha}; \{\omega_i\}_{i=1}^m \mright) = \sum_{i=1}^m \bm{\alpha}_i \omega_i$. The domain $\Lambda$ for $\bm{\alpha}$ is the probability simplex $\mleft\{ \bm{\alpha} \in \R^m : \sum_i \bm{\alpha}_i = 1, \bm{\alpha}_i \geq 0 \mright\}$.
\end{enumerate}

Our training objective is to minimize the loss $\ell$ for all network weights $\theta$ such that $ \theta =  \mathsf{P}\mleft(\bm{\alpha}, \{\omega_i\}_{i=1}^m \mright)$ for some $\bm\alpha \in \Lambda$. Recall that for input $\x$ and weights $\theta$ a neural network produces output $\hat\y = f\mleft(\x, \theta \mright)$. Given the predicted label $\hat\y$ and true label $\y$ the training loss is a scalar $\ell\mleft( \hat\y, \y\mright)$. 

If we let $\mathcal{D}$ denote the data distribution and $\mathcal{U}\mleft( \Lambda \mright)$ denote the uniform distribution over $\Lambda$, our training objective without regularization is to minimize
\begin{equation} \label{eq:objective}
    \mathbb{E}_{ (\x, \y) \sim \mathcal{D}} \mleft[ \mathbb{E}_{ \bm \alpha \sim \mathcal{U}\mleft( \Lambda \mright)} \mleft[ \ell\mleft(f\mleft(\x,  \mathsf{P}\mleft(\bm\alpha, \{\omega_i\}_{i=1}^m \mright) \mright) , \y \mright) \mright] \mright].
\end{equation}


In practice we find that achieving significant functional diversity along the subspace requires adding a regularization term with strength $\beta$ which we describe shortly. For now we proceed in the scenario where $\beta=0$. Algorithm~\ref{alg:example} is a stochastic approximation for the objective in \autoref{eq:objective}---we approximate the outer expectation with a batch of data and the inner expectation with a single sample from $\mathcal{U}\mleft( \Lambda \mright)$. 

Specifically, for each batch $(\x, \y)$ we randomly sample $\bm \alpha \sim \mathcal{U}\mleft(\Lambda\mright)$ and consider the loss
\begin{equation}
    \ell\mleft(f\mleft(\x,  \mathsf{P}\mleft(\bm\alpha, \{\omega_i\}_{i=1}^m \mright) \mright) , \y \mright).
\end{equation}
If we let $\theta = \mathsf{P}\mleft(\bm\alpha, \{\omega_i\}_{i=1}^m \mright)$ denote the single set of weights sampled from the subspace, we can calculate the gradient of each parameter $\omega_i$ as
\begin{equation}
    \frac{\partial \ell}{\partial \omega_i} = \frac{\partial \ell}{\partial \theta} \frac{ \partial \mathsf{P}\mleft(\bm\alpha, \{\omega_i\}_{i=1}^m \mright)}{\partial \omega_i}.
\end{equation}
The right hand side consists of two terms, the first of which appears in standard neural network training. The second term is computed using $\mathsf{P}$. For instance, in the case of a line, the gradient for an endpoint $\omega_1$ is
\begin{equation} \label{eq:gradient}
\frac{\partial \ell}{\partial \omega_1} = (1 - \bm\alpha) \frac{\partial \ell}{\partial \theta}.
\end{equation}
Note that the gradient estimate for each $\omega_i$ is aligned but scaled differently. As is standard for training neural networks we use SGD with momentum. In \autoref{sec:convex} we examine \autoref{eq:objective} in the simplified setting where the landscape is convex. In \autoref{sec:morereg} we approximate the inner expectation of \autoref{eq:objective} with multiple samples.

The method as described so far resembles \citet{garipov2018loss}, though we highlight some important differences. \citet{garipov2018loss} begin by independently training two neural networks and subsequently learning a connector between them, considering curves and piecewise linear functions with fixed endpoints. Our method begins by initializing the subspace parameters randomly, using the same initialization as standard training (Kaiming normal \cite{he2015delving}). The subspace is then fit in a single training run.

This contrasts significantly with standard training. For instance, when learning a simplex with $m$ endpoints we begin with $m$ random weight initializations and consider the subspace which they span. During training we move this entire subspace through the objective landscape.

\textbf{Regularization.} We have outlined a method to train high-accuracy subspaces of neural networks. However, as illustrated in \autoref{sec:reslines} (\autoref{fig:mustache-split3}), subspaces found without regularization do not contain models which achieve high accuracy when ensembled, suggesting limited functional diversity. To promote functional diversity, we want to encourage distance between the parameters $\{\omega_i\}_{i=1}^m$. 

\citet{fort2019deep} show that independently trained models have weight vectors with a cosine similarity of approximately 0, unlike models with a shared trajectory. Therefore, we encourage all pairs $\omega_j, \omega_k$ to have a cosine similarity of 0 by adding the following regularization term to the the training objective (\autoref{eq:objective}): 
\begin{align} \label{eq:reg}
\beta \cdot \mathbb{E}_{j \neq k}  \mleft[ \cos^2\mleft(\omega_j, \omega_k \mright)\mright] =  \beta \cdot \mathbb{E}_{j \neq k}  \mleft[\frac{ \mleft\langle \omega_j, \omega_k \mright\rangle^2 }{  \| \omega_j \|_{2}^2 \| \omega_k \|_{2}^2  } \mright].
\end{align}
In Algorithm \ref{alg:example} we approximate this expectation by sampling a random pair $\omega_j$, $\omega_k$ for each training batch. Unless otherwise mentioned, $\beta$ is set to a default value of 1. We do not consider $L_2$ distance since networks with batch normalization can often have weights arbitrarily scaled without changing their outputs. 

\textbf{Layerwise.} Until now our investigation has been layer agnostic---we have treated neural networks as weight vectors in $\R^n$. However, networks have structure and connectivity which are integral to their success. Accordingly, we experiment with an additional stochastic approximation to \autoref{eq:objective}. Instead of approximating the inner expectation with a single sample $\bm{\alpha} \sim \mathcal{U}\mleft( \Lambda \mright)$ we independently sample different values of $\bm{\alpha}$ for weights corresponding to different layers. In \autoref{sec:morefrankle} we extend the analysis of \citet{frankle2020linear} to this \emph{layerwise} setting.

\section{Results} \label{sec:res}

\begin{figure*}[!htbp]
    \centering
    \includegraphics[width=\textwidth]{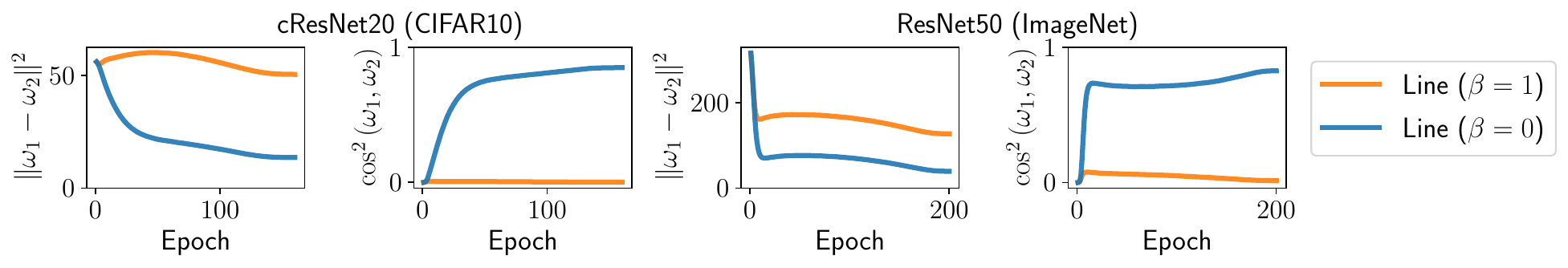}
    \vspace*{-0.75cm}
    \caption{$L_2$ distance and squared cosine similarity between endpoints $\omega_1, \omega_2$ when training a line. $\beta$ denotes the strength (scale factor) of the regularization term $\beta \cos^2\mleft(\omega_j, \omega_k \mright) = \beta \mleft\langle \omega_1, \omega_2 \mright\rangle^2 / \mleft( \| \omega_j \|_{2}^2 \| \omega_k \|_{2}^2  \mright) $ which is added to the loss to encourage large, diverse subspaces.}
    \vspace*{-0.25em}
    \label{fig:dynamics}
\end{figure*}

\begin{figure*}[h]
    \centering
    \includegraphics[width=\textwidth]{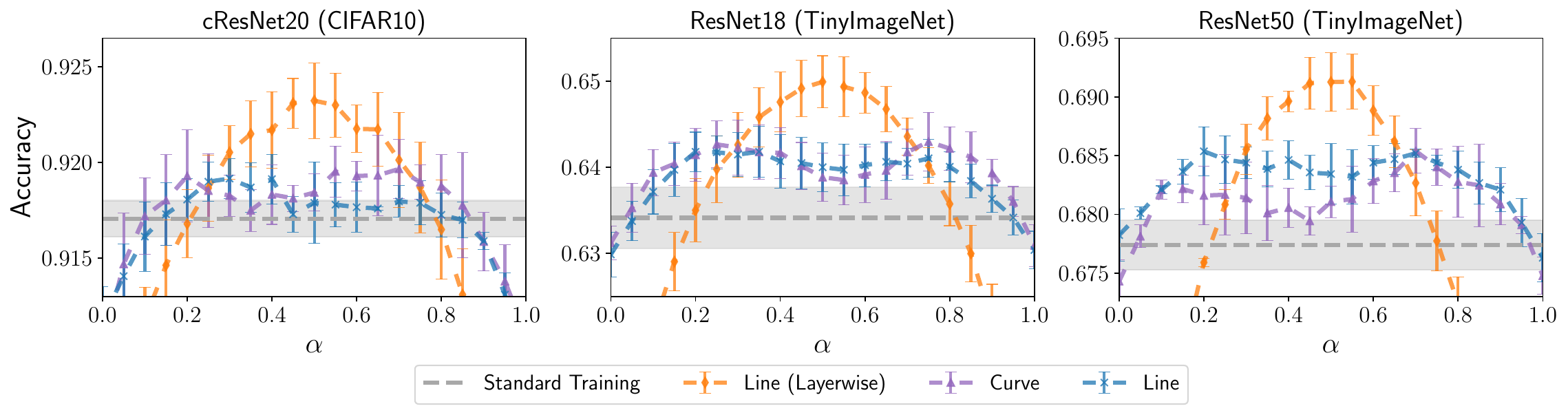}
    \vspace*{-0.75cm}
    \caption{Visualizing model accuracy along one-dimensional subspaces. The accuracy of the model at point $\alpha \in [0,1]$ along the subspace matches or exceeds standard training for a large section of the subspace (especially towards the subspace center).}
    \vspace*{-0.25em}
    \label{fig:mustache-split}
\end{figure*}
\begin{figure*}[h!]
    \centering
    \includegraphics[width=\textwidth]{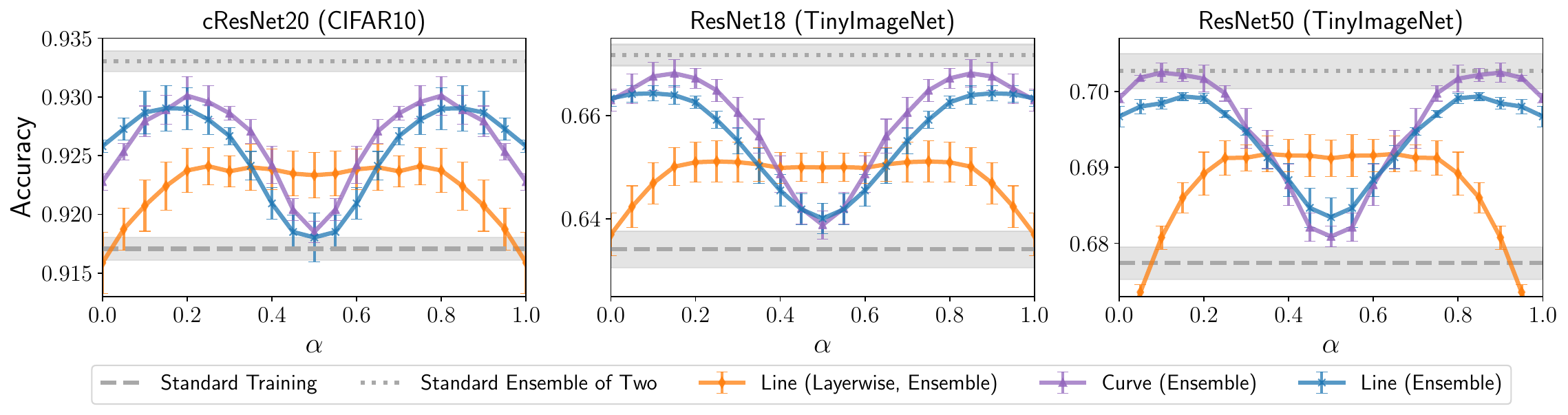}
    \vspace*{-0.75cm}
    \caption{Accuracy when two models from the subspace are ensembled---at point $\alpha$ we plot the accuracy when models $\mathsf{P(\alpha)}$ and $\mathsf{P(1-\alpha)}$ are ensembled. Performance approaches the ensemble of two independently trained networks, denoted ``Standard Ensemble of Two''.}
    \vspace*{-0.25em}
    \label{fig:mustache-split2}
\end{figure*}
\begin{figure*}[h!]
    \centering
    \includegraphics[width=\textwidth]{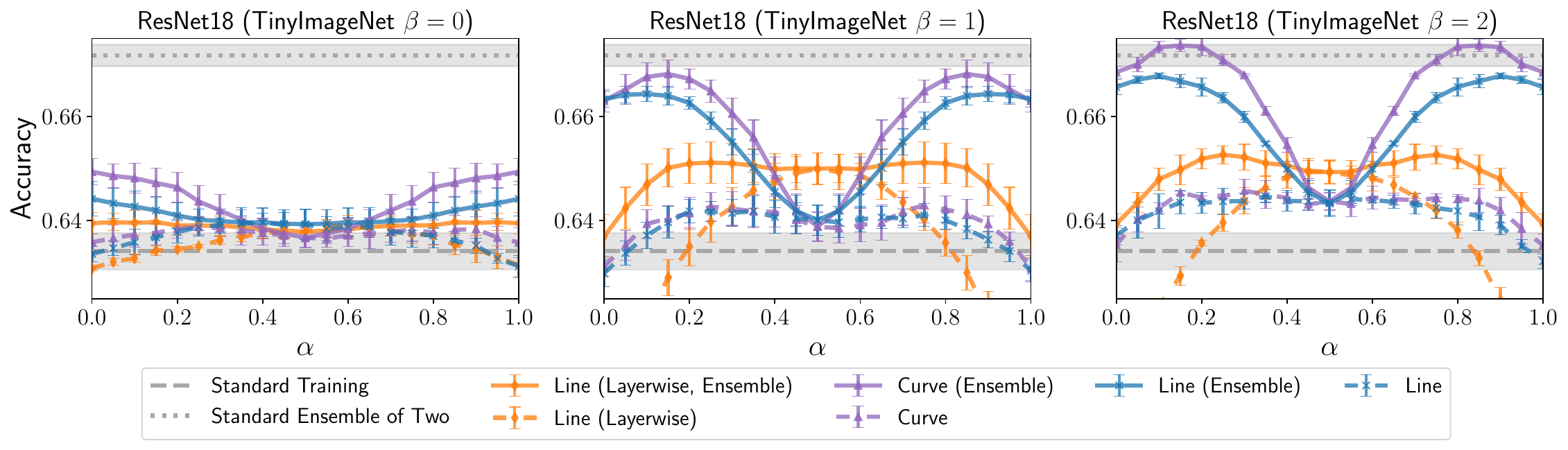}
    \vspace*{-0.75cm}
    \caption{Visualizing both model and ensemble accuracy along one-dimensional subspaces for different regularization strengths $\beta$. Regularization (\autoref{eq:reg}) tends to produce a subspace with more accurate and diverse models. Note that the visualization format of \autoref{fig:mustache-split} and \autoref{fig:mustache-split2} are combined, a technique we will use throughout the remainder of this work. For each subspace type, \textbf{(1)} accuracy of a model with weights $\mathsf{P(\alpha)}$ is shown with a dashed line and \textbf{(2)} accuracy when the output of models $\mathsf{P(\alpha)}$ and $\mathsf{P(1-\alpha)}$ are ensembled is shown with a solid line and denoted \textbf{(Ensemble)}.}
    \vspace*{-0.25em}
    \label{fig:mustache-split3}
\end{figure*}

In this section we present experimental results across benchmark datasets for image classification (CIFAR-10 \cite{krizhevsky2009learning}, Tiny-ImageNet \cite{le2015tiny}, and ImageNet \cite{deng2009imagenet}) for various residual networks \cite{he2016deep, zagoruyko2016wide}. Unless otherwise mentioned, $\beta$ (\autoref{eq:reg}) is set to a default value of 1. The CIFAR-10 \cite{krizhevsky2009learning} and Tiny-ImageNet \cite{le2015tiny} experiments follow \citet{frankle2020linear} in training for 160 epochs using SGD with learning rate 0.1, momentum 0.9, weight decay 1e-4, and batch size 128. For ImageNet we follow \citet{xie2019exploring} in changing batch size to 256 and weight decay to 5e-5. All experiments are conducted with a cosine annealing learning rate scheduler \cite{loshchilov2016sgdr} with 5 epochs of warmup and without further regularization (unless explicitly mentioned). When error bars are present the experiment is run with 3 random seeds and mean$\pm$std is shown. Additional details found in \autoref{sec:hyper}, including SWA hyperparameters and the treatment of batch norm layers (which mirror SWA \cite{izmailov2018averaging}). As discussed in \autoref{sec:runtime}, memory/FLOPs overhead is not significant as feature maps (inputs/outputs) are much larger than the number of parameters for convolutional networks. Code available at
{\small\url{https://github.com/apple/learning-subspaces}}.

\subsection{Subspace Dynamics} \label{sec:subdyn}

We begin with the following question: when training a line, how does the shape vary throughout training and how is this affected by $\beta$, the regularization coefficient? \autoref{fig:dynamics} illustrates $L_2$ distance $\| \omega_1 - \omega_2 \|_2$ and cosine similarity squared $\cos^2(\omega_1, \omega_2)$ throughout training. Recall that $\omega_1$ and $\omega_2$ denote the endpoints of the line which are initialized independently. Since a line is constructed using only two endpoints, the regularization term (\autoref{eq:reg}) simplifies to $\beta \cos^2\mleft(\omega_1, \omega_2\mright)$.

When $\beta=1$ the endpoints of a line become nearly orthogonal towards the end of training (in CIFAR10 they remain orthogonal throughout). Although $L_2$ distance isn't explicitly encouraged, it remains significant. Notably, for CIFAR10 the endpoints remain approximately as far apart throughout training as randomly initialized weights. For ResNet50 on ImageNet the $L_2$ distance between endpoints remains substantial ($\approx 127$), compared to $\approx 173$ for independently trained solutions. Note that in both cases weight decay pushes trained weights towards the origin. When $\beta=0$ there is no term encouraging separation between $\omega_1$ and $\omega_2$. However, they still remain a distance apart (13 for CIFAR10 and 40 for ImageNet). Further analysis is conducted in \autoref{sec:subdynap}, revealing that initializing $\omega_1$ and $\omega_2$ with the same shared weights has surprisingly little effect on the final cosine and $L_2$ distance.

\begin{figure*}[!htb]
    \centering
    \includegraphics[width=\textwidth]{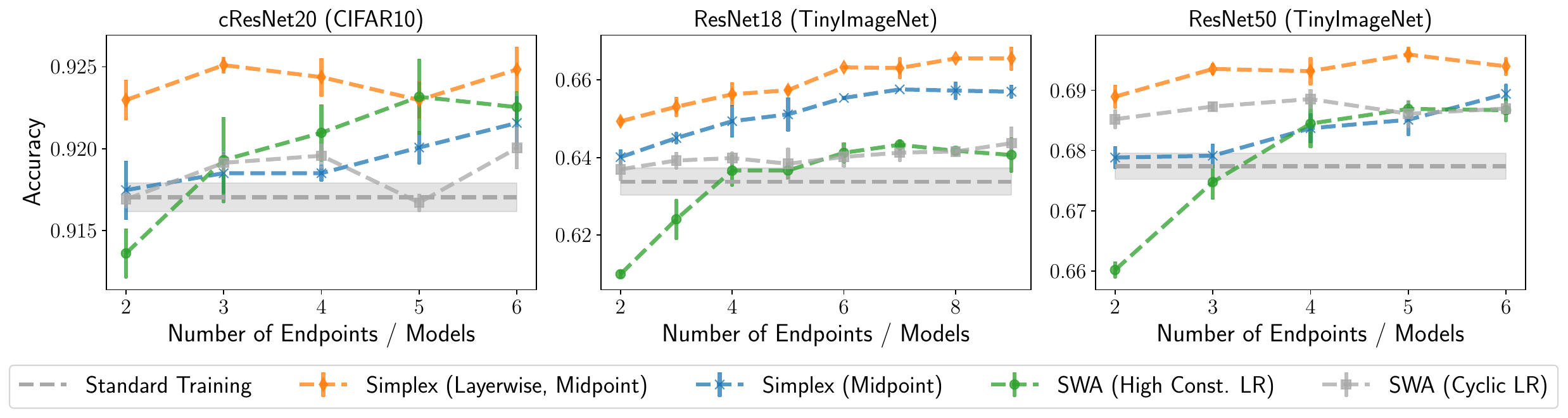}
    \vspace*{-2em}
    \caption{The model at the center of a learned simplex with $m$ endpoints improves accuracy over standard training and SWA \cite{izmailov2018averaging}. A solution towards the center of a minimum tends to be less sharp than at the periphery, which is associated with better generalization \cite{pmlr-v80-dziugaite18a}.}
    \vspace*{-1em}
    \label{fig:midpoint}
\end{figure*}

\subsection{Accuracy Along Lines and Curves} \label{sec:reslines}

Next we investigate how accuracy varies along a one-dimensional subspace. For brevity let $\mathsf{P}(\alpha)$ denote the weights at position $\alpha$ along the subspace, for $\alpha \in [0,1]$. We are interested in two quantities: (1) the accuracy of the neural network $f\mleft(\cdot, \mathsf{P}(\alpha)\mright)$ and (2) the accuracy when the outputs $f\mleft(\cdot, \mathsf{P}(\alpha)\mright)$ and $f\mleft(\cdot, \mathsf{P}(1-\alpha)\mright)$ are ensembled. Quantity (1) will determine if the subspace contains accurate solutions. Quantity (2) will demonstrate if the subspace contains diverse solutions which produce high-accuracy ensembles. 

Quantities (1) and (2) are illustrated respectively by \autoref{fig:mustache-split} and \autoref{fig:mustache-split2} 
In both \autoref{fig:mustache-split} and \autoref{fig:mustache-split2} the regularization strength $\beta$ remains at the default value of 1, while \autoref{fig:mustache-split3} provides analogous results for $\beta \in \{0,1,2\}$. Note that \textit{Layerwise} indicates that the layerwise training variant is employed (as described in \autoref{sec:meth}).

The baselines included are standard training and a standard ensemble of two independently trained networks (requiring twice as many training iterations). In \autoref{sec:more1d} we experiment with additional baselines.
There are many interesting takeaways from \autoref{fig:mustache-split}, \autoref{fig:mustache-split2}, and \autoref{fig:mustache-split3}: 
\begin{enumerate}
    \item Not only does our method find a subspace of accurate solutions, but for $\beta >0$ accuracy can improve over standard training. We believe this is because standard training solutions lie towards the periphery of a minimum \cite{izmailov2018averaging} whereas our method traverses the the minimum. Solutions at the center tend to be less sharp than at the periphery, which is associated with better generalization \cite{pmlr-v80-dziugaite18a}. These effects may be compounded by the regularization term, which leads the subspaces towards wider minima.
    \item The ensemble of two models towards the endpoints of the subspace approaches, matches, or exceeds the ensemble accuracy of two independently trained models. This is notable as the subspaces are found in only one training run.
    \item Subspaces found through the layerwise training variant have more accurate midpoints ($\alpha = 0.5$) but less accurate ensembles.
\end{enumerate}


\subsection{Performance of a Simplex Midpoint} \label{sec:simplex}

\begin{figure*}[h]
    \centering
    \includegraphics[width=\textwidth]{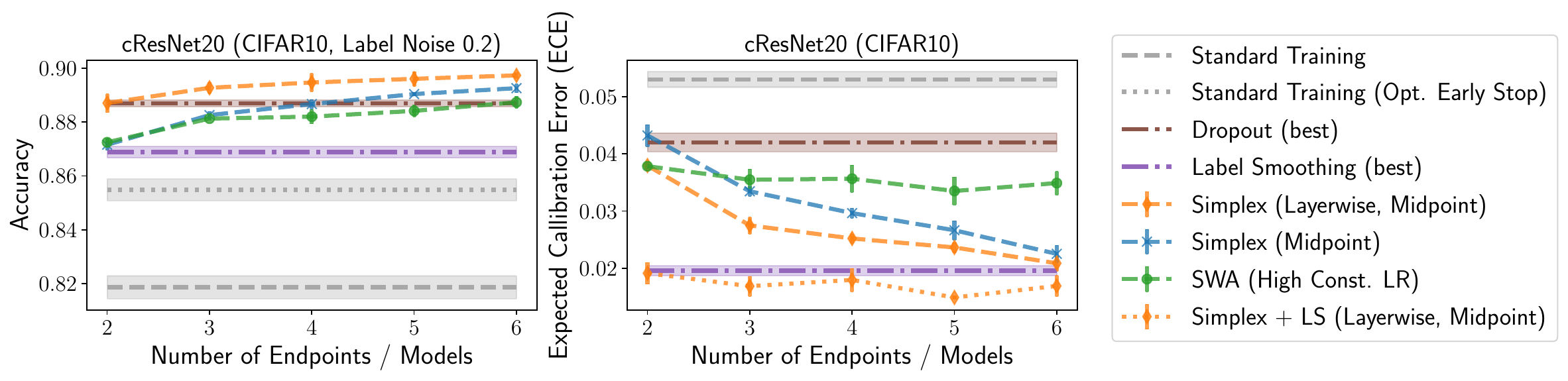}
    \vspace*{-2em}
    \caption{Using the model at the simplex center provides robustness to label noise and improved calibration. For \textit{Dropout} and \textit{Label Smoothing} we run hyperparameters $\{0.05, 0.1, 0.2, 0.4, 0.8\}$ and report the best. For Simplex + LS we add label smoothing.}
        \vspace*{-0.5em}
    \label{fig:cifarece}
\end{figure*}
\begin{figure*}[h]
    \centering
    \includegraphics[width=\textwidth]{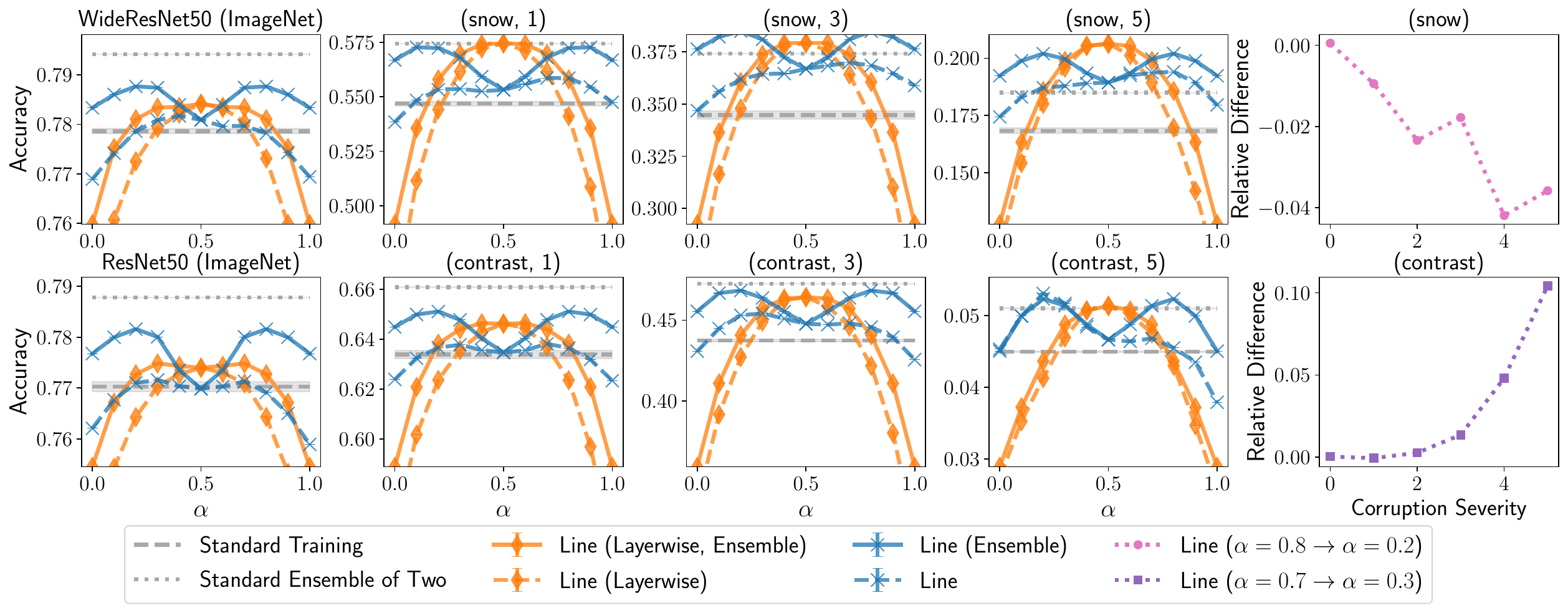}
    \vspace*{-2em}
    \caption{Accuracy along one-dimensional subspaces (with the same visualization format as \autoref{fig:mustache-split3}) tested on (left column) ImageNet \cite{deng2009imagenet} and (middle columns) ImageNet-C \cite{hendrycks2019benchmarking} for corruption types \textit{snow} and \textit{contrast} with severity levels 1, 3, and 5.  Relative difference in accuracy for two models on a line is shown in the rightmost column---models on the line with the similar performance on the clean test set exhibit varied performance on corrupted images \cite{d2020underspecification}.}
    \vspace*{-0.5em}
    \label{fig:imtest}
\end{figure*}

The previous section provided empirical evidence that the midpoint of a line (simplex with two endpoints) can outperform standard training in the same number of epochs, and hypothesized two explanations for this observation. In this section we demonstrate that this trend is amplified when considering a simplex with $m$ endpoints for $m > 2$.

\textbf{Accuracy.} The accuracy of a single model at center of a simplex is presented by \autoref{fig:midpoint}. The boost over standard training is significant, especially for TinyImageNet and higher dimensional simplexes. Recall that when training a simplex with $m$ endpoints we initialize $m$ separate networks and, for each batch, randomly sample a network in their convex hull. We then use the gradient to move this $m-1$ dimensional subspace through the objective landscape. It is not obvious that this method should converge to a high-accuracy subspace or contain high-accuracy solutions.


We compare a simplex with $m$ endpoints with SWA \cite{izmailov2018averaging} when $m$ checkpoints are saved and averaged, to maintain parity in the number of stored model parameters. For layerwise training our method outperforms or matches SWA in every case. We speculate that this may be true either because our midpoint lies closer to the minimum center than the stochastic average, or because our method finds a wider minimum then SWA. We are training a whole subspace, whereas SWA constructs a subspace after training. SWA can only travel to the widest point of the current minimum, while our method searches for a large flat minimum.

\textbf{Robustness to Label Noise; Calibration.} \autoref{fig:cifarece} demonstrates that taking the midpoint of a simplex boosts robustness to label noise and improves expected calibration error (ECE) for cResNet20 on CIFAR10. Note that CIFAR10 with label noise $c$ indicates that before training, a fraction $c$ of training data are assigned random labels (which are fixed for all methods). In addition to a SWA baseline we include optimal early stopping (the best training accuracy for standard training, before over-fitting), label smoothing \cite{muller2019does}, and dropout \cite{srivastava2014dropout}. Label smoothing and dropout have a hyperparameter for which we try values $\{0.05, 0.1,0.2,0.4,0.8\}$ and report the best result for each plot. Expected calibration error (ECE) \cite{guo2017calibration} measures if prediction confidence and accuracy are aligned. A low ECE is preferred, since models with a high ECE are overconfident when incorrect or underconfident when correct. 

\begin{figure*}
    \centering
    \includegraphics[width=\textwidth]{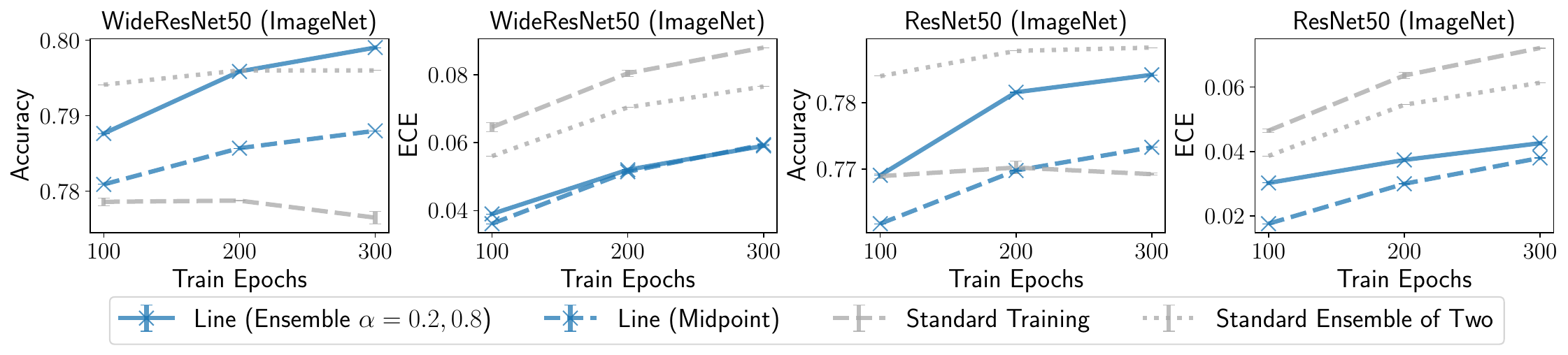}
        \vspace*{-2em}
    \caption{Accuracy and Expected Calibration Error (ECE) for the midpoint of a line trained for $\{100,200,300\}$ epochs on ImageNet. The models at the midpoint of a line are more callibrated and, when all models are trained for longer, more accurate.}
        \vspace*{-1em}
    \label{fig:imsz}
\end{figure*}

\begin{figure*}[h!]
    \centering
    \includegraphics[width=\textwidth]{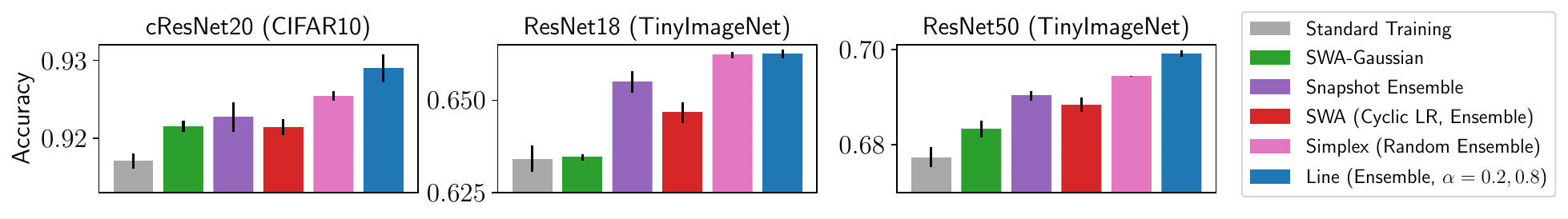}
    \vspace*{-2em}
    \caption{Ensembling 6 models drawn randomly from a 6 endpoint simplex compared with a 6 model Snapshot Ensemble \cite{huang2017snapshot}, an ensemble of 6 SWA checkpoints \cite{izmailov2018averaging}, and 6 samples from a gaussian fit to the SWA checkpoints.}
    \vspace*{-1em}
    \label{fig:imens}
\end{figure*}
\begin{figure}
    \centering
    \includegraphics[width=\columnwidth]{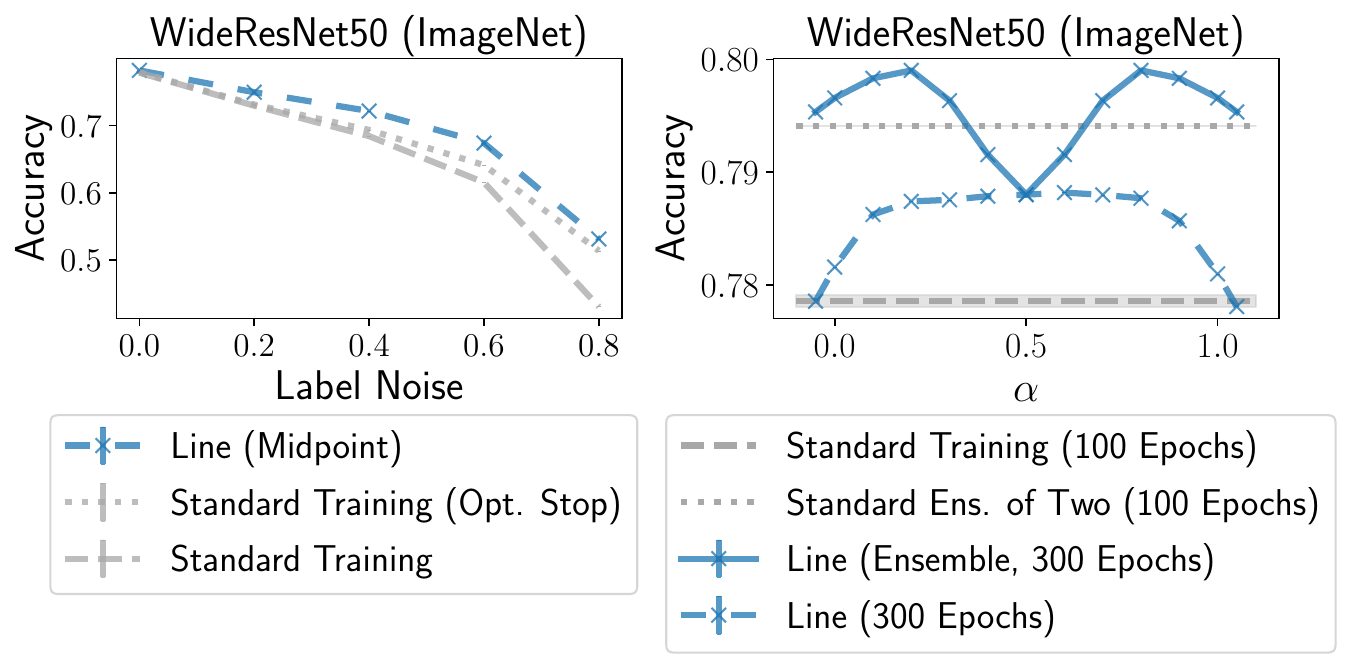}
    \vspace*{-2em}
    \caption{(left) Taking the midpoint of a line provides robustness to label noise on ImageNet compared with standard training and optimal early stopping. (right) It is possible for linearly connected models to individually attain an accuracy that is at or below standard training, while their ensemble performance is above that of standard ensembles.}
    \vspace*{-1em}
    \label{fig:imln}
\end{figure}







\subsection{ImageNet Experiments} \label{sec:large}

In this section we experiment with a larger dataset---ImageNet \cite{deng2009imagenet}---for which networks are less overparameterized. In \autoref{fig:imtest} we visualize accuracy over a line, showing both (1) the accuracy of the neural network $f\mleft(\cdot, \mathsf{P}(\alpha)\mright)$ and (2) the accuracy when the outputs of the networks $f\mleft(\cdot, \mathsf{P}(\alpha)\mright)$ and $f\mleft(\cdot, \mathsf{P}(1-\alpha)\mright)$ are ensembled. In addition to testing the network on the clean dataset (left column), we show accuracy under the \textit{snow} and \textit{contrast} dataset corruptions found in ImageNet-C \cite{hendrycks2019benchmarking}. Finally, in the right column we show the relative difference in accuracy between two models on the line. There are two interesting findings from this experiment: \textbf{(1)} it is possible to find a subspace of models, even on ImageNet, that matches or exceeds the accuracy of standard training. \textbf{(2)} Models along the line can exhibit varied robustness when faced with corrupted data.

Finding \textbf{(2)} can be examined through the lens of \emph{underspecification} in deep learning. \citet{d2020underspecification} observe that independently trained models which perform identically on the clean test set behave very differently on downstream tasks. Here we observe this behavior for models in the same linearly connected region found in a single training run. This is a promising observation in the case that a validation set exists for downstream domains. In \autoref{sec:moreimrob} we experiment with all corruptions types in ImageNet-C and demonstrate that the models we find tend to exhibit more robustness than standard training.

The WideResNet50 and ResNet50 in \autoref{fig:imtest} are respectively trained for 100 and 200 epochs (for both our method and the baseline). The smaller ResNet50 is trained for longer as, when trained for 100 epochs, the accuracy of the ResNet50 subspace falls slightly below that of standard training. However, when trained for even longer, the accuracy exceeds that of standard training. This trend is illustrated by \autoref{fig:imsz} which shows how accuracy and expected calibration error (ECE) \cite{guo2017calibration} change as a function of training epochs. The subspace midpoint is consistently more calibrated than models found through standard training.

Finally, \autoref{fig:imln} (left) demonstrates that the midpoint of a line outperforms standard training and optimal early stopping for various levels of label noise.

\subsection{Randomly Ensembling from the Subspace}

In \autoref{fig:imens} we experiment with drawing multiple models from the simplex and ensembling their predictions. We consider a simplex with 6 endpoints and draw 6 models randomly (with the same sampling strategy employed during training) and refer to the resulting ensemble as \textit{Simplex (Random Ensemble)}. We also experiment with a 6 model Snapshot Ensemble \cite{huang2017snapshot}, ensembling 6 SWA checkpoints using a cyclic learning rate (this differs slightly, but resembles FGE \cite{garipov2018loss}), and SWA-Gaussian \cite{maddox2019simple}. Additional details for the baselines are provided in \autoref{sec:hyperbase}. Surprisingly, ensembling 2 models from opposing ends of a linear subspace is still more accurate. Finally, in \autoref{sec:integral} we investigate the possibility of efficiently ensembling from a subspace without the cost.

\subsection{Is Nonlinearity Required?}

\citet{garipov2018loss, draxler2018essentially} demonstrate that there exists a nonlinear path of high accuracy between two independently trained models. Independently trained models are functionally diverse, resulting in high-performing ensembles. However, the linear path between independently trained models encounters a high loss barrier \cite{frankle2020linear, fort2020deep}. In this section we aim to provide empirical evidence which answers the following question: is this energy barrier inevitable? Is it possible for linearly connected models to individually attain an
accuracy that is at or below that of standard training, while their ensemble performance is at or above that of standard ensembles? In \autoref{fig:imln} (right) we demonstrate that, for WideResNet50 on ImageNet trained for 100 epochs, this high loss barrier is not necessary. In this one case we are concerned with existence and not training efficiency, so we find the requisite linearly connected models by training a line for 300 epochs and interpolating slightly off the line (considering $\alpha = -0.05, 1.05$).

\section{Conclusion}

We have identified and traversed large, diverse regions of the objective landscape. Instead of constructing a subspace post training, we have trained lines, curves, and simplexes of high-accuracy neural networks from scratch. However, our understanding of neural network optimization has evolved significantly in recent years and we expect this trend to continue. We anticipate that future work will continue to leverage the geometry of the objective landscape for more accurate and reliable neural networks.


\section*{Acknowledgements}
For insightful discussions, helpful suggestions, and support we thank Rosanne Liu, Jonathan Frankle, Joshua Susskind, Gabriel Ilharco Magalhães, Sarah Pratt, ML Collective, Vivek Ramanujan, Jason Yosinski, Russ Webb, Ivan Evtimov, and Hessam Bagherinezhad. We acknowledge Ludwig Schmidt for correcting Definitions 1 and 2 which previously measured average accuracy instead of worst case. MW acknowledges Apple for providing internship support.


\bibliography{ref}
\bibliographystyle{icml2021}

\cleardoublepage
\appendix

\begin{figure*}[!htbp]
    \centering
    \includegraphics[width=\textwidth]{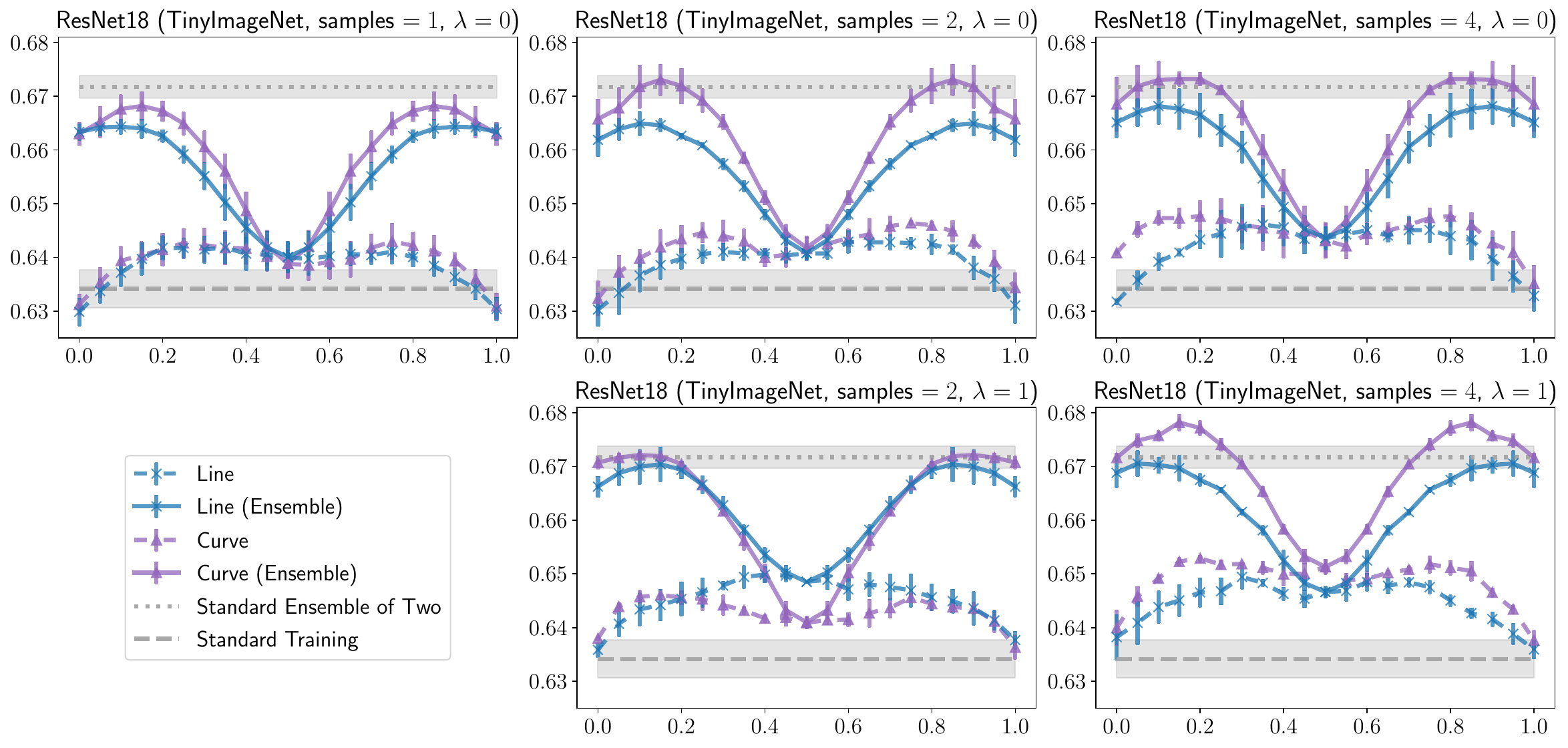}
    \vspace*{-0.75cm}
    \caption{Model and ensemble accuracy along one-dimensional subspaces. For each subspace type, \textbf{(1)} accuracy of a model with weights $\mathsf{P(\alpha)}$ is shown with a dashed line and \textbf{(2)} accuracy when the output of models $\mathsf{P(\alpha)}$ and $\mathsf{P(1-\alpha)}$ are ensembled is shown with a solid line and denoted \textbf{(Ensemble)}. The number of samples of $\alpha$ used to approximate the inner expectation of \autoref{eq:objective} is given by \textit{samples} while $\lambda$ denotes the strength of the feature similarity regularization (\autoref{sec:morereg}). Both \textit{samples} $> 1$ and $\lambda > 0$ tend to improve accuracy for both lines and curves.}
    \label{fig:mustache2}
\end{figure*}

\section{Convex Setting} \label{sec:convex}


In this section we consider the case where the loss is convex, and we show the optimization problem remains convex when learning the subspace parameters.

Let $\bm{\omega} = (\omega_1,...,\omega_m)$ denote the parameters used to construct the subspace. A simplified version of our objective is given by
\begin{equation}
    h\mleft( \bm{\omega} \mright) \triangleq \mathbb{E}_{\bm{\alpha}\sim \mathcal{U}\mleft(\Lambda\mright)} \mleft[ \ell \mleft( \mathsf{P}\mleft(\bm{\alpha}, \bm{\omega} \mright)  \mright) \mright]
\end{equation}
This objective is simplified from \autoref{eq:objective} as we have removed the dependence on the training data and neural network---the loss $\ell$ is given parameters $\theta \in \R^n$ and returns a positive scalar.

We note that in each of the subspaces we learn---lines, curves, and simplexes---$\mathsf{P}(\bm{\alpha}, \bm{\omega})$ is linear with respect to $\bm{\omega}$. 

\begin{proposition}
If $\ell : \R^n \rightarrow \R$ is convex and $\mathsf{P}$ is linear with respect to $\bm{\omega}$ then $h$ is convex with respect to $\bm{\omega}$.
\begin{proof}
For two sets of parameters $\bm{\omega}$ and $\overline{\bm{\omega}}$ and $\lambda \in [0,1]$,
\begin{align}
    &h\mleft((1-\lambda) \bm{\omega} + \lambda \overline{\bm{\omega}} \mright) \\
    &= \mathbb{E}_{\bm{\alpha}} \mleft[ \ell\mleft(
    \mathsf{P}\mleft(\bm{\alpha}, (1-\lambda) \bm{\omega} + \lambda \overline{\bm{\omega}} \mright) \mright) \mright] \\
    &= \mathbb{E}_{\bm{\alpha}} \mleft[ \ell\mleft(
    (1-\lambda)  \mathsf{P}\mleft(\bm{\alpha}, \bm{\omega}\mright) + \lambda \mathsf{P}\mleft(\bm{\alpha}, \overline{\bm{\omega}} \mright)\mright) \mright] \label{eq:linp} \\
    &\leq \mathbb{E}_{\bm{\alpha}} \mleft[ (1-\lambda)  \ell\mleft(
     \mathsf{P}\mleft(\bm{\alpha}, \bm{\omega}\mright) \mright) + \lambda \ell\mleft( \mathsf{P}\mleft(\bm{\alpha}, \overline{\bm{\omega}} \mright)\mright) \mright] \label{eq:cvxl} \\
    &= (1-\lambda)  \mathbb{E}_{\bm{\alpha}} \mleft[  \ell\mleft(
     \mathsf{P}\mleft(\bm{\alpha}, \bm{\omega}\mright) \mright)\mright] + \lambda \mathbb{E}_{\bm{\alpha}} \mleft[ \ell\mleft( \mathsf{P}\mleft(\bm{\alpha}, \overline{\bm{\omega}} \mright)\mright) \mright] \\
    &= (1-\lambda) h\mleft( \bm{\omega} \mright) + \lambda h\mleft(\overline{\bm{\omega}} \mright),
\end{align}
where \autoref{eq:linp} and \autoref{eq:cvxl} respectively follow from the linearity of $\mathsf{P}$ (in $\bm{\omega}$) and convexity of $\ell$.
\end{proof}
\end{proposition}

\section{Additional Samples and Feature Similarity Regularization} \label{sec:morereg}

In Algorithm~\ref{alg:example} we approximate the inner expectation of \autoref{eq:objective} using a single sample. In this section we approximate the expectation with multiple samples, leading to an improvement in accuracy along the subspace and of the ensemble.
When approximating the expectation with $s$ samples we split the batch of size $b$ into $s$ groups of size $b / s$ and sample independent values of $\alpha \sim \mathcal{U}([0,1])$ for each. Results for $s = \{1,2,4\}$ are shown in the first row of \autoref{fig:mustache2}.

Using multiple samples allows us to experiment with additional regularization to enable functional diversity. We can directly encourage models from different parts of the subspace to have orthogonal features. We experiment with regularization of this form, which we call feature similarity regularization, in the second row of \autoref{fig:mustache2}. For each batch we pick $j,k$ randomly from $\{1,...,s\}$, where $s$ is the number of samples. Let $\alpha_j$ and $\alpha_k$ denote samples $j$ and $k$ from $\mathcal{U}([0,1])$ and $\phi_j$, $\phi_k$ denote the features obtained using models $\mathsf{P}(\alpha_j)$ and $\mathsf{P}(\alpha_k)$. The feature similarity regularization term is then given by
\begin{align}
    \lambda |\alpha_j - \alpha_k | \cos^2(\phi_j, \phi_k )
\end{align}
where the features $\phi$ are taken from the output of the penultimate layer and $\cos(\phi_j, \phi_k)$ is cosine similarity. The term $ |\alpha_j - \alpha_k | $ allows for more feature similarity when models are close together on the subspace. Results for feature similarity regularization are shown in the bottom row of \autoref{fig:mustache2}.

\section{Integrating over Subspaces} \label{sec:integral}

Is there a subspace from which you can efficiently ensemble \textit{all} models? We believe this is not possible for the subspaces of general neural networks $f$ we learn in this paper. However, this does become possible when considering a specific form for $f$.

\begin{figure}
    \centering
    \includegraphics[width=\columnwidth]{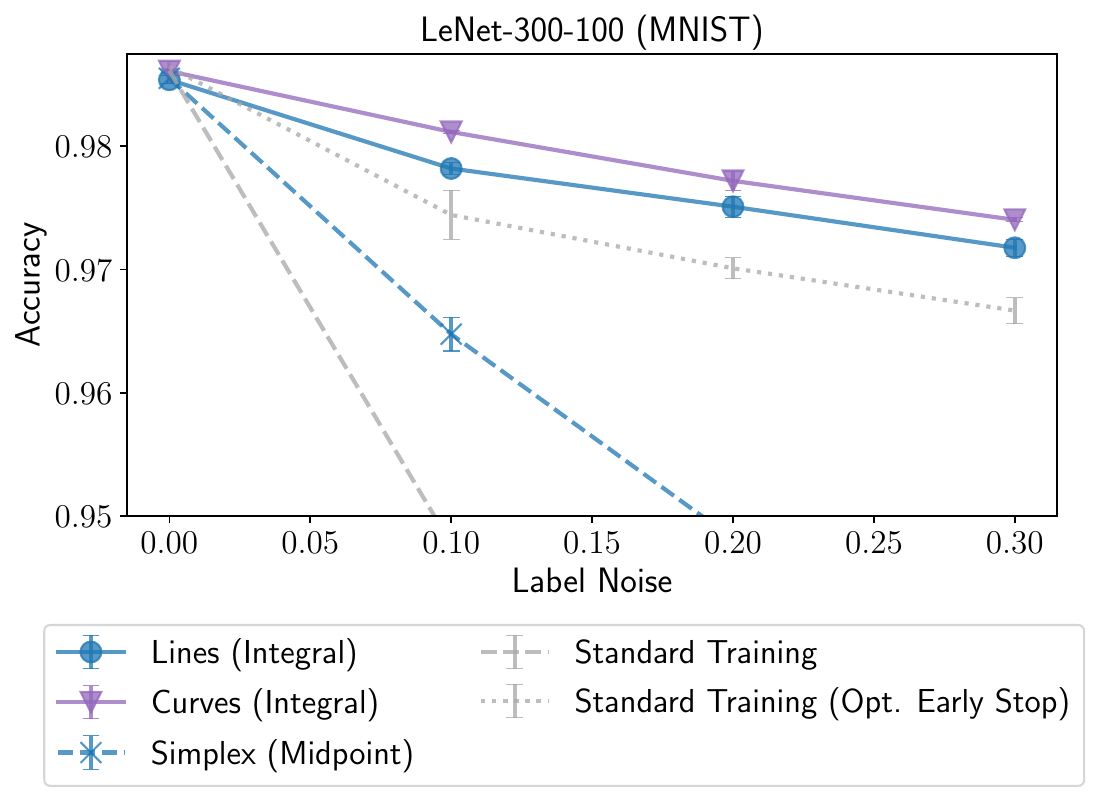}
    \caption{Learning subspaces of functions with efficient closed-form continuous ensembles (\autoref{eq:integral2}). Since the functional form is restricted, these ``Integral'' solutions only provide an accuracy boost for nonzero label noise.}
    \label{fig:integral}
\end{figure}

Consider $\mathsf{P} : [0,1] \rightarrow \R^n$ which defines a one-dimensional subspace of weights. In this section we investigate a mechanism for ensembling the output of all networks along the subspace---a closed-form expression for
\begin{align} \label{eq:integral}
    \hat y(\x) = \int_0^1 f\mleft(\x, \mathsf{P}(\alpha) \mright) d\alpha.
\end{align}
For a particular class of functions $f$, \autoref{eq:integral} admits a straightforward solution. Consider
\begin{align}
    f(\x, \mathsf{P}(\alpha)) = g(\x, \mathsf{P}(0)) + \frac{dg\mleft(\x, \mathsf{P}(\alpha)) \mright) }{d\alpha} .
\end{align}
for which
\begin{align} \label{eq:integral2}
    \int_0^1 f\mleft(\x, \mathsf{P}(\alpha) \mright) d\alpha = g(\x, \mathsf{P}(1)).
\end{align}
The function $g$ can be any learned neural network. To train $f$ (\textit{i.e.} to learn $g$) we approximate the derivative by finite difference during training. For each training batch $(\x, \y)$ we sample $\alpha$ uniformly from $[0,1]$ and compute outputs
\begin{align}
    f(\x, \mathsf{P}(\alpha)) = g(\x, \mathsf{P}(0)) + \frac{g(\x, \mathsf{P}(\alpha + \epsilon)) - g(\x, \mathsf{P}(\alpha))}{\epsilon}.
\end{align}
During evaluation we then return $g(\x, \mathsf{P}(1))$ which corresponds to the ensemble of all networks $f(\x, \mathsf{P}(\alpha))$ (\autoref{eq:integral2}). As shown in \autoref{fig:integral}, we experiment with this model on MNIST \cite{lecun1998mnist} using $\epsilon = 0.1$. We use \textbf{Integral} to refer to the model described in this section. Recall that a label noise level of $c$ denotes that a fraction $c$ of the training data is assigned random and fixed labels before training. Since we are restricting the form of $f$, the accuracy does not differ significantly from standard training when there is no label noise. However, as label noise increases the integral solution outperforms other models.

\section{Additional Experimental Details} \label{sec:hyper}

\subsection{Models and Training Details.} \label{sec:models}

For CIFAR10 experiments we use the ResNet20 model (referred to as cResNet20) which may be found at 
{\small\url{https://github.com/facebookresearch/open_lth}}. For TinyImageNet we use the ResNet$\{18,50\}$ models which were used by \citet{tanaka2020pruning} in their TinyImageNet experiments. Finally, the ImageNet models are from PyTorch \cite{NEURIPS2019_9015}. We use PyTorch 1.6 and Python 3.7. All models are trained on a single GPU except for the ImageNet models which are trained on 4 GPUs. Standard data augmentations are used---random crop and horizontal flip. To sample uniformly from the $m-1$ dimensional probability simplex we sample $m$ random variables from the exponential distribution then normalize so that the sum is 1.
\begin{figure*}[!htbp]
    \centering
    \includegraphics[width=\textwidth]{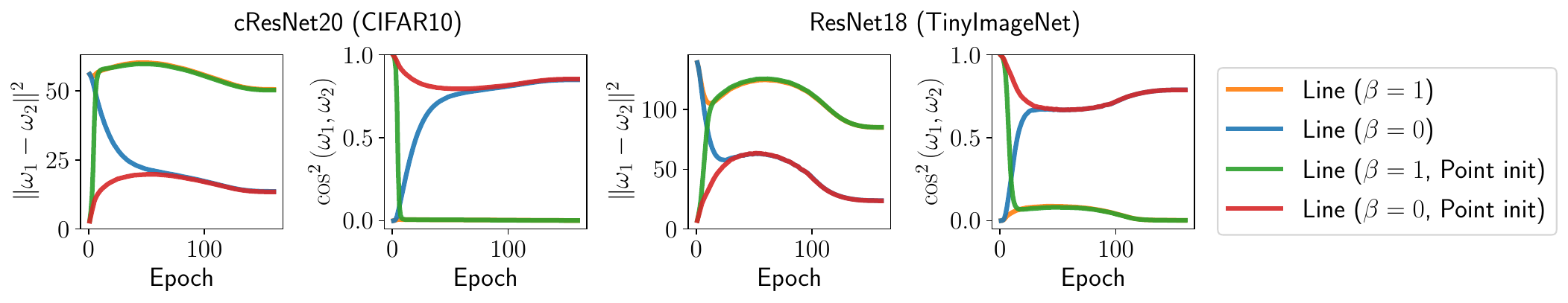}
    \vspace*{-1.0cm}
    \caption{$L_2$ distance and squared cosine similarity between endpoints $\omega_1, \omega_2$ when training a line. For ``Point init'' the endpoints of the line were initialized with the same shared weight values.} 
    \label{fig:dynamicsv4}
\end{figure*}
\begin{figure*}[!htbp]
    \centering
    \includegraphics[width=\textwidth]{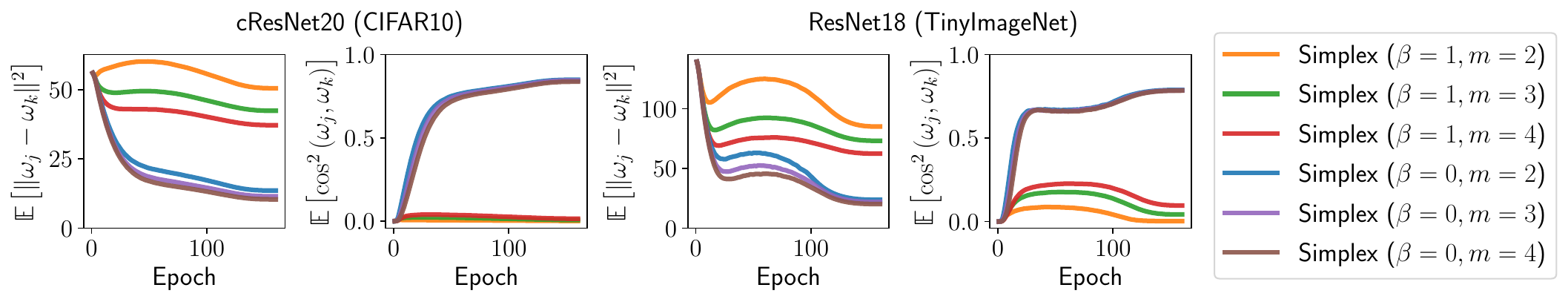}
    \vspace*{-1.0cm}
    \caption{Average $L_2$ distance and squared cosine similarity between endpoints $\omega_j, \omega_k$ when training an $m$ endpoint simplex with regularization strength $\beta$ (\autoref{eq:reg}).} 
    \label{fig:dynamicsv6}
\end{figure*}

\subsection{Computation} \label{sec:runtime}

Consider a convolutional layer with kernel size $\kappa \times \kappa$, input size $(b, c_1, w_1, h_1)$, and output size $(b, c_2, w_2, h_2)$. The number of parameters is $p = c_1 c_2 \kappa^2$ while the number of FLOPs in standard training is $M = b p  w_2 h_2$. The algorithm we present requires $O(p (m-1))$ additional FLOPs to update the subspace's network weights, where $m$ is the number of parameters used to construct the subspace. This overhead is minimal with respect to $M$ (since $b, w_2, h_2$ tend to be large, $b$ alone is over 100). The only storage overhead comes from storing multiple copies of the model parameters $O(p (m-1))$, which is not significant compared to buffers stored for the backward pass, of size $O(bc_2w_2h_2)$ \cite{mxnet}. This is especially true for lines, curves, and low dimensional simplexes which constitute the majority of our experiments. No additional storage is required for computing the gradient, since the gradient updates to each endpoint are re-scaled versions of the same tensor (except the gradient of the regularization term, which has no dependence on the input data and can be computed after the initial buffers are freed).



\subsection{Batch Normalization} \label{sec:bn}

In many cases batch norm \cite{ioffe2015batch} parameters require different treatment then other network weights. In standard training the batch norm scale parameter is initialized to be a vector of ones, and often remains mainly positive. Accordingly, cosine distance is likely the wrong distance metric to compare batch norm parameters. Moreover, the number of batch norm parameters is very small with respect to the total number of weights. Accordingly, in \autoref{fig:dynamics} and \autoref{fig:dynamicsv4} we do not take batch norm parameters into account when considering cosine or $L_2$ distance. Moreover, in Algorithm~\ref{alg:example} we do not take batch norm parameters into account when computing the regularization term (\autoref{eq:reg}).

Although we train batch norm parameters which lie on a line, curve, or simplex, batch norm layers also track a running mean and variance. Since these are not learned parameters, we follow \citet{izmailov2018averaging, maddox2019simple} and recompute these statistics using training data. For instance, when evaluating the model at the midpoint of the simplex we first compute the running mean and variance with a pass through the training data before evaluating on the test set. For group norm \cite{wu2018group} or layer norm \cite{ba2016layer} this would not be an issue, although these methods tend to achieve lower accuracy than batch norm in the settings we consider.

\subsection{Baseline Hyperparameters} \label{sec:hyperbase}

We implement all baselines with the same hyperparameters described in \autoref{sec:res} whenever possible. However, some baselines have additional hyperparameters. For SWA \cite{izmailov2018averaging} we use the default values from {\small{\url{https://github.com/timgaripov/swa}}}---SWA LR of 0.05 and begin saving checkpoints 40 epochs before training ends (75\% of the way through). For experiments with SWA throughout this wok we use either a cyclic (denoted \textit{Cyclic LR}) or high constant (denoted \textit{High Const. LR})  learning rate for the late phase of training and provide results for the best or both. For SWA-Guassian we construct the Gaussian using 6 saved SWA checkpoints.

Additionally, we tried using the regularization term (\autoref{eq:reg}) to encourage diversity among the SWA checkpoints but did not succeed in improving performance.

\begin{figure*}
    \centering
    \includegraphics[width=\textwidth]{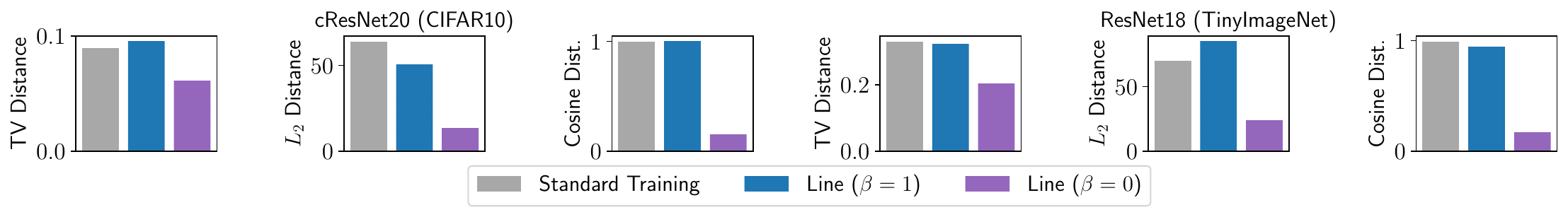}
        \vspace*{-0.8cm}
    \caption{Comparing the statistics of the models which lie at the endpoints of a learned line with two independently trained models. We compare total variation (TV) distance between the outputs and \{$L_2$, Cosine\} distance between the weights.}
    \label{fig:stats}
\end{figure*}

\begin{figure*}[h!]
    \centering
    \includegraphics[width=\textwidth]{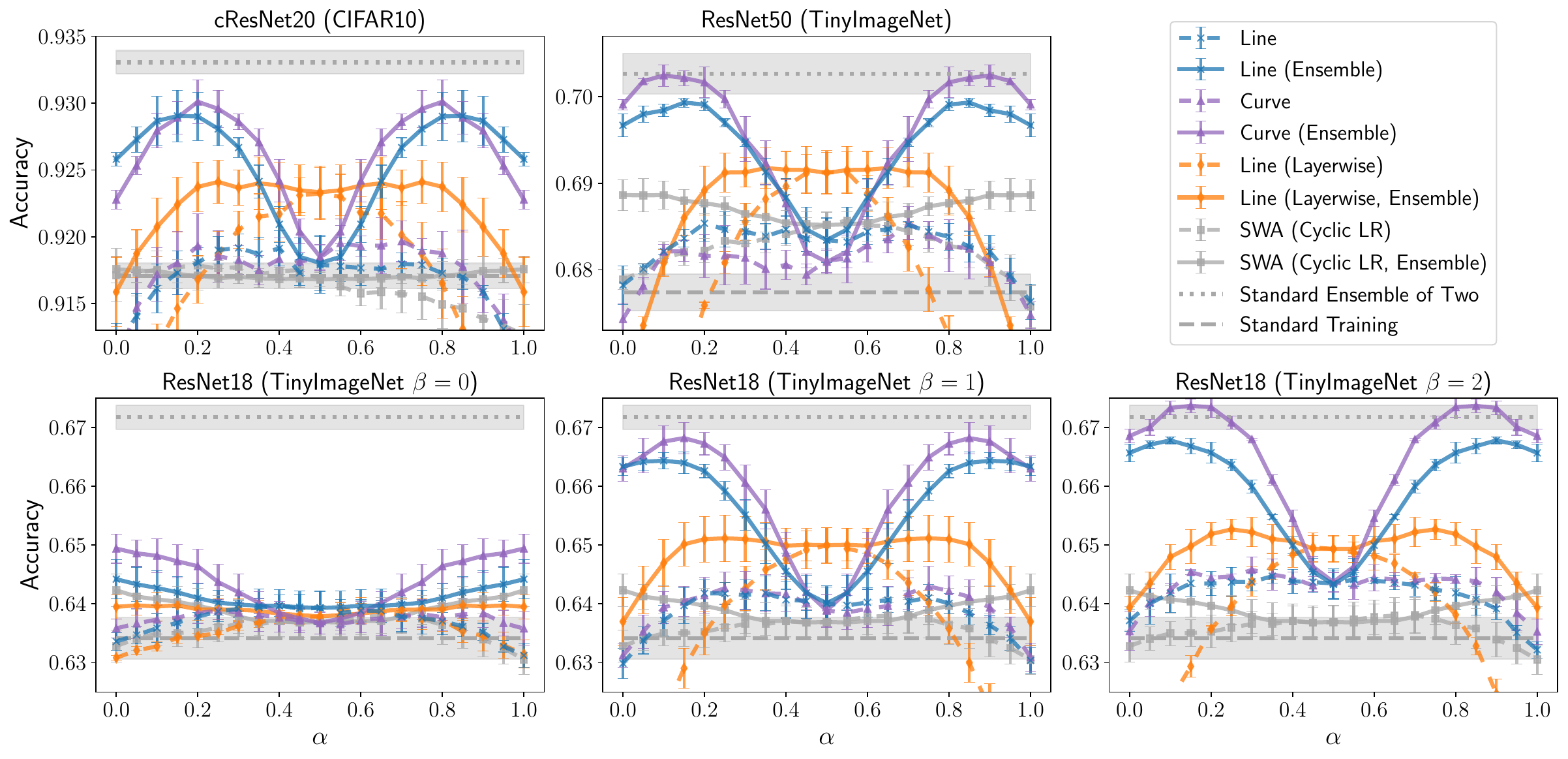}
    \vspace*{-0.75cm}
    \caption{Model and ensemble accuracy along one-dimensional subspaces. For each subspace type, \textbf{(1)} accuracy of a model with weights $\mathsf{P(\alpha)}$ is shown with a dashed line and \textbf{(2)} accuracy when the output of models $\mathsf{P(\alpha)}$ and $\mathsf{P(1-\alpha)}$ are ensembled is shown with a solid line and denoted \textbf{(Ensemble)}. Note that quantity (2) is symmetric about 0.5 at which point it also intersects with quantity (1). ``Standard Ensemble of Two'' is the ensemble accuracy of two independently trained networks. For SWA we save only two checkpoints and consider the subspace formed by interpolating between them.}
    \vspace*{-0.5em}
    \label{fig:mustache}
\end{figure*}

\begin{figure*}[h!]
    \centering
    \includegraphics[width=\textwidth]{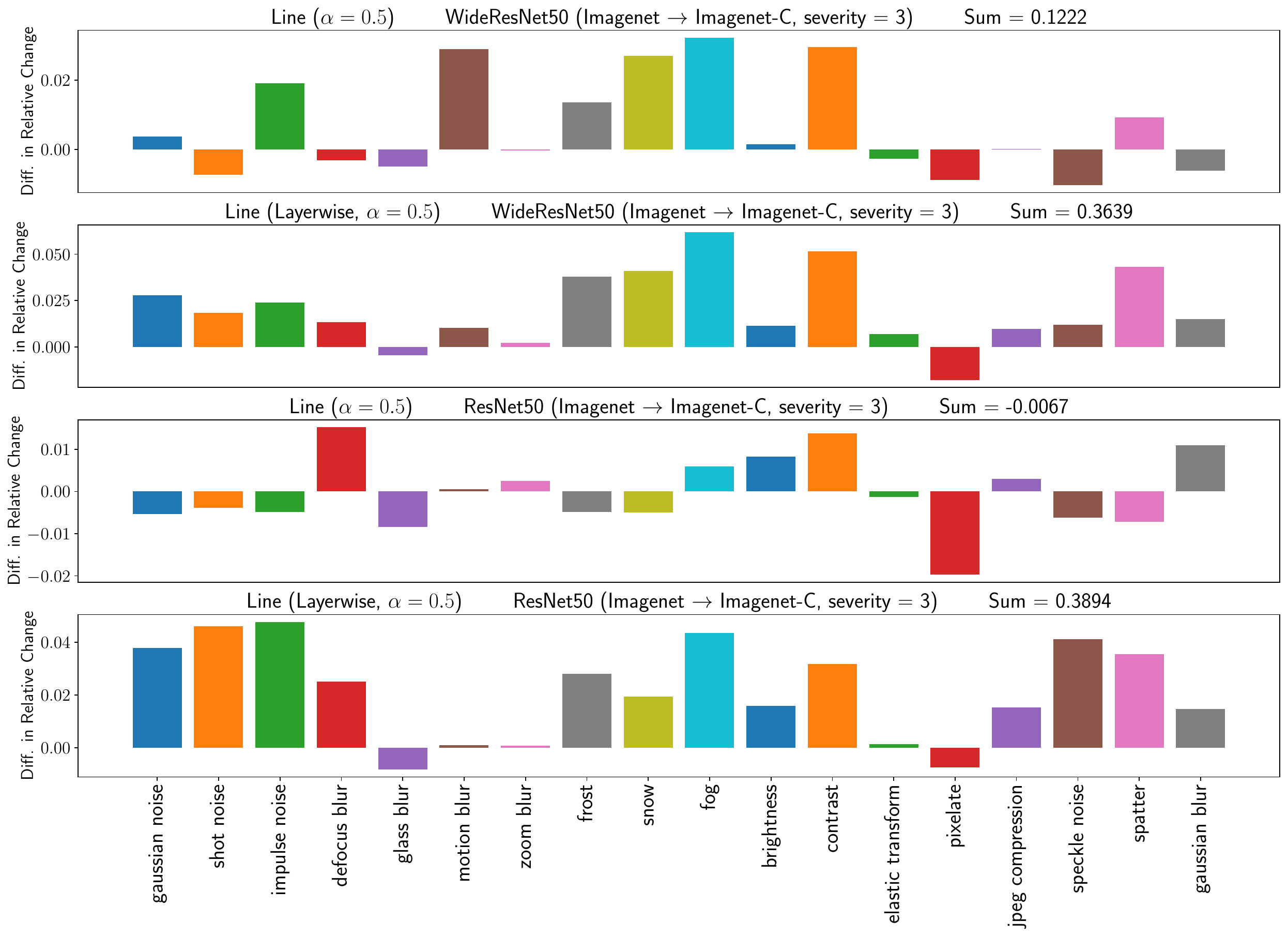}
    \caption{Comparing the relative change in accuracy when tested on corrupted data. Comparison is between the midpoint of a line and standard training. A positive bar indicates that the midpoint of the line has relatively less of a drop in accuracy from clean to corrupted data in ImageNet-C. \cite{hendrycks2019benchmarking}.  See text (\autoref{sec:moreimrob}) for details.}
    \label{fig:bar}
\end{figure*}

\begin{figure*}[h!]
    \centering
    \includegraphics[width=\textwidth]{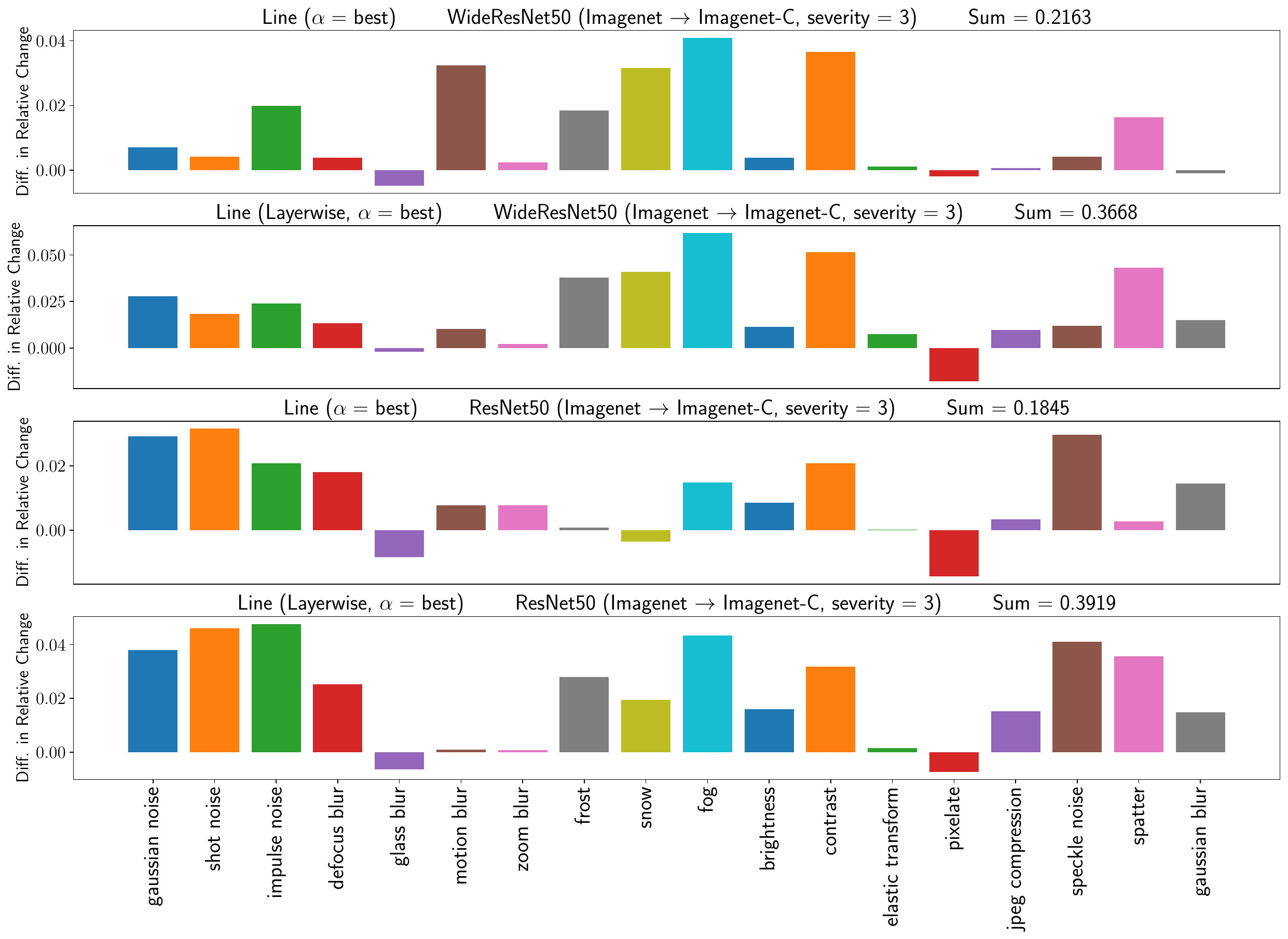}
    \caption{Comparing the relative change in accuracy when tested on corrupted data. Comparison is between the best model on the line and standard training. A positive bar indicates that there exists a model on the line with relatively less of a drop in accuracy from clean to corrupted data in ImageNet-C. This result demonstrates that there exists a model on the line which performs well, but does not indicate how to find this model. See text (\autoref{sec:moreimrob}) for details.}
    \label{fig:barbest}
\end{figure*}

\section{Further Subspace Dynamics}\label{sec:subdynap}

This section extends the results from \autoref{sec:subdyn} which examine the shape of subspace throughout training. \autoref{fig:dynamicsv4} illustrates that initializing the endpoints of the line with the same shared initialization (``point init'') has little effect on the dynamics. After a few epochs of training, any discrepency between ``{point init}'' and standard initialization nearly disappears. In \autoref{fig:dynamicsv6} we examine the average $L_2$ distance and squared cosine similarity between endpoints when training simplexes. The same general trends hold, but the average distance between endpoints decreases with the number of endpoints $m$. Since a new random pair of endpoints is sampled for each batch in Algorithm~\ref{alg:example}, closeness between each individual pair is penalized less for larger $m$. 
Finally, in \autoref{fig:stats} we compare the endpoints of a line with two independently trained models in terms of $L_2$ distance, cosine distance, and total variation (TV) distance. For two networks with outputs $\p_1$ and $\p_2$ the TV distance is given by $\frac{1}{2}\|\p_1 - \p_2 \|_1$ and is averaged over all examples in the test set. As expected, $\beta=1$ produces lines with more distant and functionally diverse endpoints.

\section{Additional Baselines for One-Dimensional Subspaces}\label{sec:more1d}

In \autoref{fig:mustache} we augment the experiments from \autoref{sec:reslines} the additional baseline of SWA (described in \autoref{sec:prelim}) with a cyclic learning rate scheduler. For experiments with SWA \cite{izmailov2018averaging} throughout this wok we use either a cyclic (denoted \textit{Cyclic LR}) or high constant (denoted \textit{High Const. LR})  learning rate for the late phase of training and provide results for the best or both. In the case of \autoref{fig:mustache}, where we save only two SWA checkpoints and interpolate between, cyclic performs better as the high constant scheduler does not find checkpoints which match standard training accuracy. Additional details on baseline hyperparameters are provided in \autoref{sec:hyper}.

\section{Additional ImageNet-C Robustness Experiments} \label{sec:moreimrob}

In this section we test the models trained on ImageNet (\autoref{fig:imtest}, \autoref{sec:large}) across all image corruptions in the ImageNet-C dataset \cite{hendrycks2019benchmarking}. We consider the relative change in accuracy when models are evaluated on corrupted images. For a model with accuracy $a$ on the clean set and $b$ on the corrupted images, the relative change in accuracy is $(b-a)/a$. The relative change in accuracy (which we refer to as \textit{relative change}) is chosen because performance on the clean test set can act as a confounder \cite{taori2020measuring}. The experiments are conducted with a corruption severity of 3.

\autoref{fig:bar} illustrates the difference in \textit{relative change} between the midpoint of the line and a model found through standard training. A positive value indicates that the midpoint of the line has a relatively less severe drop in accuracy when faced with corrupted data. Although performance on different corruption types is varied, the midpoint models we find tend to exhibit more robustness---WideResNet50 (layerwise) outperforms standard training on all but two corruption types. 

However, this evaluation considers only the midpoint of the line, ignoring that we have trained family of models. In \autoref{fig:barbest}, we compare the best-performing model on the line (over $\alpha \in \{0, 0.1, ..., 0.9, 1.0\}$, in terms of the relative change in accuracy) with standard training. This setting is not realistic as $\alpha$ is tuned on the test set, however it is a positive sign for the case when a validation set exists for the corrupted data of interest. A single training run can capture a family of models, and each can be tested on a validation set for the downstream domain. 


\begin{figure*}
    \centering
    \includegraphics[width=\textwidth]{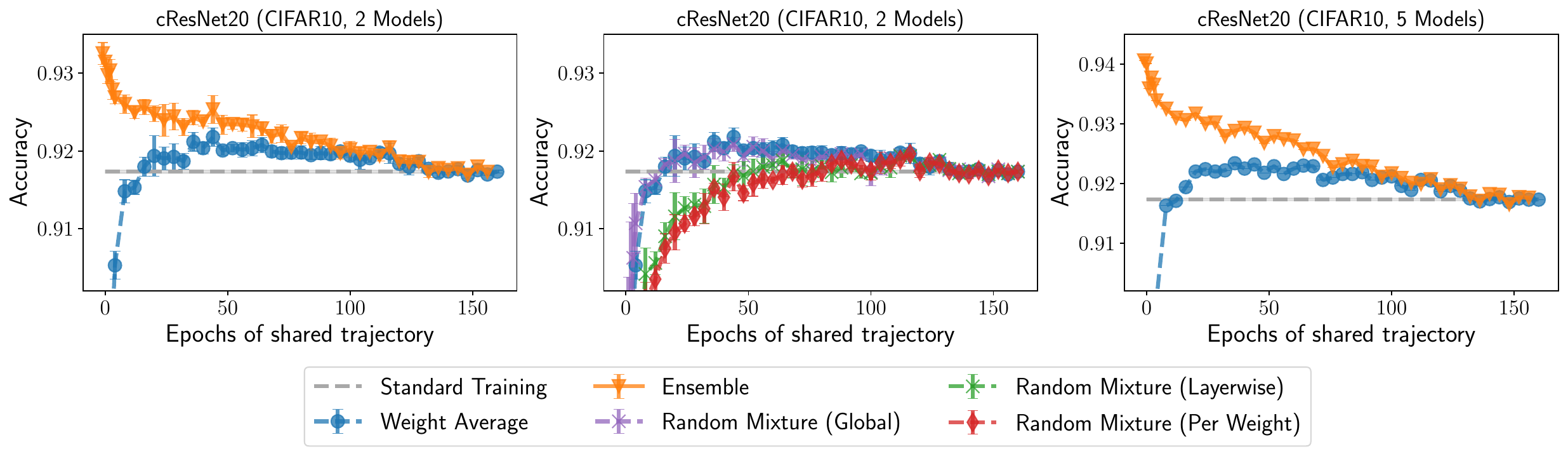}
    \caption{(left) Reproducing the instability analysis of \cite{frankle2020linear}---providing the accuracy of the weight space ensemble and output space ensemble of two models with $k$ epochs of shared trajectory. (middle) Considering random interpolations between models with $k$ epochs of shared trajectory. Interpolations are global, per-layer, and per-weight. (right) Extending the instability analysis result to 5 models.}
    \label{fig:wa}
\end{figure*}
\section{Further Analysis of \citet{frankle2020linear}} \label{sec:morefrankle}

Recall from \autoref{sec:prelim}, Observation 4 that \citet{frankle2020linear} consider the scenario where two networks branch off after $k$ epochs of the trajectory are shared. In other words, they consider $\ws_k = \mathsf{Train}^{0 \rightarrow k}\mleft(\ws_0, \xi \mright)$ and let $\ws^i_{k\rightarrow T} = \mathsf{Train}^{k \rightarrow T}\mleft(\ws_k, \xi_i \mright)$ for $i \in \{1,2\}$. In \autoref{fig:wa} (left) we reproduce results from \citet{frankle2020linear}, demonstrating that for very small $k$, the weight average of $\theta^1_{k\rightarrow T}$ and $\theta^2_{k\rightarrow T}$ matches the accuracy standard training accuracy. Note that the weight average refers to the accuracy of model $f\mleft( \cdot, \frac{1}{2}\mleft(\ws^1_{k\rightarrow T} + \ws^2_{k\rightarrow T}\mright) \mright)$ and the ensemble refers to the accuracy of model $\frac{1}{2}\mleft(f\mleft( \cdot, \ws^1_{k\rightarrow T}  \mright) + f\mleft( \cdot, \ws^2_{k\rightarrow T}  \mright)\mright)$. For moderate $k$, the weight average exceeds standard training as a result of Observation 5 (\autoref{sec:prelim}).

\autoref{fig:wa} (right) demonstrates that these results hold when considering weight and output space ensembles of 5 models which all share $k$ epochs of trajectory. Finally, in \autoref{fig:wa} (middle) we consider random interpolations at different scales. Random Mixture (Global) is given by
\begin{align}
    \mathbb{E}_{\alpha \sim \mathcal{U}([0,1])}\mleft[ \mathsf{Acc}\mleft( (1- \alpha)\theta^1_{k\rightarrow T} + \alpha \theta^2_{k\rightarrow T}.
    \mright) \mright]
\end{align}
For Random Mixture (Layerwise) we sample different coefficients $\alpha$ for each layer, and for Random Mixture (Per-weight) we sample different $\alpha$ for all weights in the network. The latter corresponds to a hyper-rectangle with corners $\theta^1_{k\rightarrow T}$ and $\theta^2_{k\rightarrow T}$. When $k$ is at least half of thhe training epochs all models on this $n$ dimensional hyper-rectangle match standard training accuracy.



\end{document}